\documentclass[twoside]{article}

\usepackage[accepted]{aistats2023}
\usepackage[colorlinks=true,citecolor=blue]{hyperref}

\usepackage{times}
\usepackage{epsfig}
\usepackage{graphicx}
\usepackage{amsmath}
\usepackage{amssymb}

\usepackage{times}  
\usepackage{helvet}  
\usepackage{courier}
\usepackage{tikz}
\usepackage{comment}
\usepackage[utf8]{inputenc} 
\usepackage[T1]{fontenc}    
\usepackage{url}            
\usepackage{booktabs}       
\usepackage{amsfonts}       
\usepackage{nicefrac}      
\usepackage{microtype}     
\usepackage{algorithm}
\usepackage{algpseudocode}
\usepackage{array}
\usepackage{bm}
\usepackage{multirow}
\usepackage{color}
\usepackage{newfloat}
\usepackage{listings}
\usepackage{subcaption}
\usepackage{adjustbox}
\usepackage{makecell}

\usepackage{caption}
\usepackage{subcaption}

\begin{document}

\runningtitle{Generative Radiance Field with Latent Space Energy-Based Model for 3D-Aware Disentangled Image Representation}

\twocolumn[

\aistatstitle{Likelihood-Based Generative Radiance Field with Latent Space Energy-Based Model for 3D-Aware Disentangled Image Representation}

\aistatsauthor{ Yaxuan Zhu \And Jianwen Xie, Ping Li }

\aistatsaddress{Department of Statistics\\University of California - UCLA\\  8125 Math Sciences Bldg, 
Los Angeles, CA, USA\\ yaxuanzhu@ucla.edu \And Cognitive Computing Lab\\Baidu Research\\10800 NE 8th St. Bellevue WA 98004, USA\\\{jianwen.kenny, pingli98\}@gmail.com} ]

\begin{abstract}
\footnote{The work of Yaxun Zhu was conducted as a research intern at Cognitive Compuitng Lab, Baidu Research -- Bellevue, WA, USA}We propose the NeRF-LEBM, a likelihood-based top-down 3D-aware 2D image generative model that incorporates 3D representation via Neural Radiance Fields (NeRF) and 2D imaging process via differentiable volume rendering.  
The model represents an image as a rendering process from 3D object to 2D image and is conditioned on some latent variables that account for object characteristics and are assumed to follow informative trainable energy-based prior models. We propose two likelihood-based learning frameworks to train the NeRF-LEBM: (i) maximum likelihood estimation with Markov chain Monte Carlo-based inference and (ii) variational inference with the reparameterization trick. We study our models in the scenarios with both known and unknown camera poses. Experiments on several benchmark datasets demonstrate that the NeRF-LEBM can infer 3D object structures from 2D images, generate 2D images with novel views and objects, learn from incomplete 2D images, and learn from 2D images with known or unknown camera poses.
\end{abstract}

\section{Introduction}\label{sec:introduction}

\subsection{Motivation}
Towards the goal of 3D-aware image synthesis, existing methods  generate 3D representations of objects either in a voxel-based format~\cite{zhu2018visual} or via intermediate 3D features~\cite{alhaija2018geometric}, and then use the differentiable rendering operation to render the generated 3D object into 2D views. However, the voxel-based 3D representation is discrete and memory-inefficient so that the methods are limited to generating low-quality and low-resolution images of objects. Notably, the Neural Radiance Field (NeRF)~\cite{mildenhall2022nerf} has become a new type of 3D representation of objects and shown impressive results for new view synthesis. It represents a continuous 3D scene or object by a mapping function parameterized by a neural net, which takes as input a 3D location and a viewing direction and outputs the values of color and density. The visualization of the 3D object can be achieved through generating different views of images by querying the mapping function at each specific 3D location and viewing direction, followed by volume rendering operation~\cite{kajiya1984ray} to produce image pixel intensities. In general, each NeRF function can only represent one single object and need to be trained from multiple views of images of that object. By generalizing the original NeRF function to a conditional version that involves latent variables that account for the appearance and shape of the object, GRAF~\cite{schwarz2020graf} builds a 2D image generator based on the conditional NeRF and trains the generator for 3D-aware controllable image synthesis via adversarial learning. The NeRF-VAE~\cite{kosiorek2021nerf} proposes to train the NeRF-based generator via variational inference, where the bottom-up inference network allows the inference of 3D structures of objects in unseen testing images. Both GRAF and NeRF-VAE assumes the object-specific latent variables to follow simple and non-informative Gaussian distributions. As a likelihood-based model, the NeRF-VAE can only handle training images with known camera poses because of the difficulty of inference of the unknown camera pose for each observed image. The GRAF, which is a non-likelihood-based generative model, can easily learn from images with unknown camera poses because its adversarial learning scheme does not need to deal with inference. Recently, we have witnessed the rapid advance of adversarial NeRF-based generative models, however, the progress in developing likelihood-based NeRF-based generative models has been lagging behind. Conceptually, likelihood-based generative models have many advantages, e.g., stable learning process without a mode collapse issue, capability of inferring latent variables from training and testing examples, and capability of learning from incomplete data via unsupervised learning. Thus, this paper aims at pushing forward the progress of likelihood-based generative radiance field.         

To be specific, by leveraging the NeRF-based image generator and the latent space energy-based models (LEBMs) ~\cite{pang2020learning}, this paper proposes the NeRF-LEBM, a novel likelihood-based 3D-aware generative model for 2D images. It builds energy-based models (EBMs) on the latent space of the NeRF-based generator~\cite{schwarz2020graf}. The latent space EBMs are treated as informative prior distributions. We can follow the empirical Bayes, and train the EBM priors and the NeRF-based generator simultaneously from observed data. The trainable EBM priors over latent variables (appearance and shape of the object) allow sampling novel objects from the model and rendering images with arbitrary viewpoints, as well as improve the capacity of the latent spaces and the expressivity of the NeRF-based generator. Suppose there is a set of 2D training images presenting multiple objects with various appearance, shapes and viewpoints. We first study the scenario of~\cite{kosiorek2021nerf}, in which the viewpoint of each image is known. We propose to train the models by maximum likelihood estimation (MLE) with Markov chain Monte Carlo~\cite{barbu2020monte} (MCMC)-based inference, in which no  extra assisting network is required. At each iteration, the learning algorithm runs MCMC sampling of the latent variables from the EBM priors and the posteriors. The update of the EBM priors is based on the samples from the prior and the posterior distributions, while the update of the generator is based on the samples from the posteriors and the observed data. Furthermore, for efficient training and inference, we also propose to use the amortized inference to train the NeRF-LEBM as an alternative. Lastly, we do not assume the camera pose of each image is given and treat it as latent variables that follows a uniform prior distribution. We propose to use the von Mises-Fisher (vMF)~\cite{davidson2018hyperspherical} distribution to approximate the posterior of the camera pose in our amortized inference framework. Our experiments show that the proposed likelihood-based generative model can not only synthesize images with new objects and arbitrary viewpoints but also learn meaningful disentangled representation of images in scenarios of both known and unknown camera poses. The model can even learn from incomplete 2D training images for control generation and 3D aware inference.  Our paper makes the following contributions: 
\begin{enumerate}
 
\item We propose a novel NeRF-based 2D generative model, with a  trainable energy-based latent space, for 3D-aware image synthesis and disentangled representation.

\item We propose to train the model by MLE with MCMC-based inference, which does not rely on separate networks and is more principled and statistically rigours than adversarial learning and variational learning. 
\item We propose to train the model via amortized inference by recruiting inference networks. Due to the use of EBM priors, our algorithm is different from the learning framework of the NeRF-VAE.
\item We are the first to solve the problem of inferring unknown camera poses in variational inference framework through a novel posterior and prior setting.  

\item We conduct extensive experiments to test the efficiency, effectiveness and performance of the proposed NeRF-LFBM model and learning algorithms.    
\end{enumerate}

\subsection{Related work}

\paragraph{3D-aware image synthesis} Prior works study controllable image generation by adopting 3D data as supervision~\cite{wang2016generative,zhu2018visual}  or 3D information  as input~\cite{alhaija2018geometric,oechsle2019texture}. Several works~\cite{kato2018neural,henzler2019escaping,navaneet2019differ,liu2019soft} build discriminative mapping functions from 2D images to 3D shapes, followed by differentiable rendering to project the 3D generated objects back to images for computing reconstruction errors on image domain. Unlike the aforementioned reconstruction-based frameworks, several recent works, such as GRAF~\cite{schwarz2020graf}, GIRAFFE~\cite{niemeyer2021giraffe}, pi-GAN~\cite{chan2021pi},  and NeRF-VAE~\cite{kosiorek2021nerf}, build 2D generative models with NeRF function and differentiable rendering and assume unobserved object-specific variables to follow known Gaussian prior distributions. They are trained by adversarial learning~\cite{goodfellow2014generative} or variational inference~\cite{kingma2014auto}. Our model is also a NeRF-based generative model, but assumes latent object-specific variables to follow informative prior distributions parameterized by energy-based models~\cite{xie2016theory}. We propose to train NeRF-based generator and EBM priors simultaneously by likelihood-based learning with either MCMC or amortized~inference.    

\paragraph{Energy-based models} 
Recently, with the striking expressive power of modern deep networks, deep data space EBMs~\cite{xie2016theory} have shown impressive performance in modeling distributions of different types of high-dimensional data, e.g., images~\cite{nijkamp2019learning,zhao2021learning,gao2020learning, XieZL21,ZhengXL21,XieZLL22}, videos~\cite{xie2017synthesizing, xie2021learning_pami}, 3D volumetric shapes~\cite{xie2018learning, xie2022generative}, and unordered point clouds~\cite{xie2021generative}. Besides, deep latent space EBMs~\cite{pang2020learning}, which stand on generator networks and serve as prior distributions of the latent variables, have proven to be effective in learning expressive latent spaces for text~\cite{pang2021latent, yu2022latentdiffusion}, image~\cite{pang2020learning},  trajectory generation~\cite{pang2021trajectory}, and saliency prediction~\cite{zhang2021learning}. In our paper, we build EBMs on the latent spaces of a NeRF-based generator to serve as prior distributions of object appearance and shape for camera pose-conditioned~image~generation. 

\paragraph{MCMC inference} Our model is also related to the theme of training deep latent variable models with MCMC inference \cite{HanLZW17,XieGZZW19, ZhuXLE19, XieGZZW20, ZhangXB20, NijkampP0ZZW20, AnXL21, zhang2021learning, abs-2301-09300}. Different from the above works, our paper is the first to study the MCMC inference in a NeRF-based generator with two EBM priors. 


\section{Background}
\subsection{Neural Radiance Field} 
A continuous scene can be represented by a Neural Radiance Field (NeRF)~\cite{mildenhall2022nerf}, which is a mapping function $f_{\theta}$ whose input is a 3D location $\mathbf{x} \in \mathbb{R}^3$ and a 3D unit vector as viewing direction $\mathbf{d} \in \mathbb{R}^3$, and whose output is an RGB color value $\mathbf{c} \in \mathbb{R}^3$ and a volume density $\sigma$. Formally, $f_{\theta}: (\textbf{x},\textbf{d}) \rightarrow (\textbf{c},\sigma)$,

where $f_{\theta}$ is a neural network with parameters $\theta$. Given a fixed camera pose, to render a 2D image from the NeRF representation $f_{\theta}$, we can follow the  classical volume rendering method~\cite{kajiya1984ray} to calculate the color of each pixel $\textbf{v} \in \mathbb{R}^2$ in the 2D image. The color of the pixel is determined by the color and volume density values of all points along the camera ray $r$ that goes through that pixel $\textbf{v}$. In practice, we can follow~\cite{mildenhall2022nerf} and sample $M$ points $\{\textbf{x}^r_i\}_{i=1}^M$from the near to far bounds along the camera ray $r$ and obtain a set of corresponding colors and densities  $\{(\textbf{c}^r_i,\sigma^r_i)\}_{i=1}^M$ by $f_{\theta}$, and then we compute the color $C(r)$ for the camera ray $r$ by
\begin{eqnarray}
\begin{aligned}
C(r) = \sum_{i=1}^{M} T^r_i (1-\exp(-\sigma^r_i \delta^r_i))\textbf{c}_i^r,
\label{eq:render}
\end{aligned}
\end{eqnarray}
where $\delta^r_i=||\textbf{x}^r_{i+1}-\textbf{x}^r_i||_2$ is the distance between adjacent sample points, and $T_i^r=\exp(-\sum_{j=1}^{i-1} \sigma^r_j \delta^r_j)$ is the accumulated transmittance along the ray from the 1st point to the $i$-th point, i.e., the probability that the ray travels from $\textbf{x}^r_1$ to $\textbf{x}^r_i$ without being blocked. 
To render the whole image $\textbf{I}$, we need to compute the color for the ray that corresponds to each pixel $\textbf{v}$ in the image. Let $r(\textbf{v})$ be the camera ray corresponding to the pixel $\textbf{v}$, and the rendered image is given by  $\textbf{I}(\textbf{v})=C(r(\textbf{v})), \textbf{v} \in \mathcal{D}$, where  $\mathcal{D}$ is the image domain.

\subsection{Conditional Neural Radiance Field}
The original NeRF function $f_{\theta}$ is a 3D representation of a single scene or object. To generalize the NeRF to represent different scenes or objects,~\cite{schwarz2020graf} proposes the conditional NeRF function
\begin{eqnarray}
\begin{aligned}
g_{\theta}: (\textbf{x},\textbf{d}, \textbf{z}^s, \textbf{z}^a) \rightarrow (\textbf{c},\sigma),
\label{eq:cnerf}
\end{aligned}
\end{eqnarray}

which is conditioned on object-specific variables, $\textbf{z}^a$ and $\textbf{z}^s$, corresponding to object appearance and shape respectively. It can be further decomposed into (i) $g^1_{\theta_1}:   (\textbf{x},\textbf{z}^s) \rightarrow \textbf{h}$, (ii) $g^2_{\theta_2}:  (\textbf{h},\textbf{d},\textbf{z}^a) \rightarrow \textbf{c}$, and (iii) $g^3_{\theta_3}: \textbf{h} \rightarrow \sigma$ to show the dependency among the input variables in the design of $g(\theta)$. 

\section{Proposed framework}
\subsection{NeRF-based 2D generator with EBM priors}

We are interested in learning a 3D-aware generative model of 2D images, with the purposes of controllable image synthesis and disentangled image representation. We build a top-down 2D image generator based on a conditional NeRF structure for the intrinsic 3D representation of the object in an image. Let $\textbf{z}^a$ and $\textbf{z}^s$ be the latent variables that define the shape and the appearance of an object, respectively. $\textbf{z}^a$ and $\textbf{z}^s$ are assumed to be independent. They together specify an object. Let $\boldsymbol{\xi}$ be the camera pose. 
The generator $G_{\theta}$ consists of an object-conditioned NeRF function $g_{\theta}$ as shown in Eq.~(\ref{eq:cnerf}) and a differentiable rendering function as shown in Eq.~(\ref{eq:render}). $\theta$ are trainable parameters of the generator. Given an object specified by $(\textbf{z}^a, \textbf{z}^s)$, the generator takes the camera pose $\boldsymbol{\xi}$ as input and outputs an image by using the NeRF $g_{\theta}$ to render an image from the pose $\boldsymbol{\xi}$ with the render operation in Eq.~(\ref{eq:render}). Given a dataset of 2D images of different objects captured from different viewing angles (i.e., different camera pose), in which the camera pose of each image is provided. We assume each image is generated by following the generative process defined by $G_{\theta}$ and each of the latent variables $(\textbf{z}^a,\textbf{z}^s)$ is assumed to follow an informative prior distribution that is defined by a trainable energy-based model (EBM). Specifically, the proposed 3D-aware image-based generative model is given by the following deep latent variable model
\begin{eqnarray}
\begin{aligned}
&\textbf{I}=G_{\theta}(\textbf{z}^a,\textbf{z}^s,\boldsymbol{\xi})+\epsilon,\\
&\epsilon \sim \mathcal{N}(0,\sigma_{\epsilon}^2I),\\
&\textbf{z}^a\sim p_{\alpha_a}(\textbf{z}^a),\\
&\textbf{z}^s\sim p_{\alpha_s}(\textbf{z}^s), 
\label{eq:generator}
\end{aligned}
\end{eqnarray}
where $\epsilon$ is the observation residual following a Gaussian distribution $\mathcal{N}(0,\sigma_{\epsilon}^2I)$ with a known standard deviation $\sigma_{\epsilon}$, and $I$ denotes the identity matrix. Both $p_{\alpha_a}(\textbf{z}^a)$ and $p_{\alpha_s}(\textbf{z}^s)$ are modeled by EBMs
\begin{align}
p_{\alpha_a}(\textbf{z}^a)&=\frac{1}{Z(\alpha_a)}\exp[-U_{\alpha_a}(\textbf{z}^a)]q_0(\textbf{z}^a),\\  
p_{\alpha_s}(\textbf{z}^s)&=\frac{1}{Z(\alpha_s)} \exp[-U_{\alpha_s}(\textbf{z}^s)]q_0(\textbf{z}^s),
\label{eq:ebm}
\end{align}
which are in the form of exponential tilting of a Gaussian reference distribution $q_0 \sim \mathcal{N}(0,\sigma^2 I)$. (Note that $q_0$ could be a uniform reference distribution.) $U_{\alpha_a}(\textbf{z}^a)$ and $U_{\alpha_s}(\textbf{z}^s)$ are called energy functions, both of which are parameterized by multilayer perceptrons (MLPs) with trainable parameters  $\alpha_a$ and $\alpha_s$, respectively. The energy function takes the corresponding latent variables as input and outputs a scalar as energy. Besides, $Z(\alpha_a)=\int \exp[-U_{\alpha_a}(\textbf{z}^a)]q_0(\textbf{z}^a)d \textbf{z}^a$ and $Z(\alpha_s)=\int \exp[-U_{\alpha_s}(\textbf{z}^s)]q_0(\textbf{z}^s) d \textbf{z}^s$ are intractable normalizing constants. Although $q_0(\textbf{z}^a)$ and $q_0(\textbf{z}^s)$ are Gaussian distributions, $p_{\alpha_{a}}(\textbf{z}^a)$ and $p_{\alpha_{z}}(\textbf{x}^s)$ are non-Gaussian priors, where $\alpha_{a}$ and $\alpha_{s}$ are learned from the data together with the parameters $\theta$ of the generator $G_{\theta}$. 

\subsection{Learning with MCMC-based inference}
For convenience of notation, let $\beta=(\theta, \alpha_a, \alpha_s)$ and $\alpha=(\alpha_a, \alpha_s)$. Given a set of 2D images with known camera poses, i.e., $\{(\textbf{I}_i,\boldsymbol{\xi}_i), i=1,...,n\}$, we can train $\beta$ by maximizing the observed-data log-likelihood function defined~as
\begin{eqnarray}
\begin{aligned}
    &L(\beta)=\frac{1}{n}\sum_{i=1}^{n} \log p_{\beta}(\textbf{I}_i|\boldsymbol{\xi}_i) \\  
    &=\frac{1}{n}\sum_{i=1}^{n} \log \left[\int p_{\alpha}(\textbf{z}_i^a,\textbf{z}_i^s)p_{\theta}(\textbf{I}_i|\textbf{z}_i^a,\textbf{z}_i^s,\boldsymbol{\xi}_i)d \textbf{z}_i^a d\textbf{z}_i^s \right]\\
    &=\frac{1}{n}\sum_{i=1}^{n} \log \left[\int p_{\alpha_a}(\textbf{z}_i^a) p_{\alpha_s}(\textbf{z}_i^s)p_{\theta}(\textbf{I}_i|\textbf{z}_i^a,\textbf{z}_i^s,\boldsymbol{\xi}_i)d \textbf{z}_i^a d\textbf{z}_i^s \right],\nonumber 
    \label{eq:likelihood}
\end{aligned}
\end{eqnarray}
where $p_{\alpha}(\textbf{z}^a, \textbf{z}^s)=p_{\alpha_a}(\textbf{z}^a)p_{\alpha_s}(\textbf{z}^s)$  because $\textbf{z}^a$ and $\textbf{z}^s$ are statistically independent, and the latent variables are integrated out in the complete-data log-likelihood.  According to the Law of Large Number, maximizing the likelihood $L(\beta)$ is approximately equivalent to minimizing the Kullback-Leibler (KL) divergence between model $p_{\beta}(\textbf{I}|\boldsymbol{\xi})$ and data distribution  $p_{\text{data}}(\textbf{I}|\boldsymbol{\xi})$ if the number $n$ of training examples is very large. The gradient of $L(\beta)$ is calculated based~on
\begin{align}
    \nabla_{\beta} &\log p_{\beta}(\textbf{I}|\boldsymbol{\xi})
    = \mathbb{E}_{p_{\beta}(\textbf{z}^a,\textbf{z}^s|\textbf{I},\boldsymbol{\xi})} \left[\nabla_{\beta} \log p_{\beta}(\textbf{I},\textbf{z}^a,\textbf{z}^s|\boldsymbol{\xi}) \right] \notag\\
    =&\mathbb{E}_{p_{\beta}(\textbf{z}^a,\textbf{z}^s|\textbf{I},\boldsymbol{\xi})}[\nabla_{\beta} \log p_{\alpha_a}(\textbf{z}^a) + \nabla_{\beta} \log p_{\alpha_s}(\textbf{z}^s) \nonumber\\
    &+ \nabla_{\beta} \log p_{\theta}(\textbf{I}|\textbf{z}^a,\textbf{z}^s,\boldsymbol{\xi}) ], \label{eq:gradient}
\end{align}
which can be further decomposed into three parts, i.e., the gradients for the EBM prior of object appearance $\alpha_a$
\begin{align}
    &\mathbb{E}_{p_{\beta}(\textbf{z}^a, \textbf{z}^s|\textbf{I}, \boldsymbol{\xi})}\left[ \nabla_{\beta} \log p_{\alpha_a}(\textbf{z}^a)\right] \label{eq:prior_gradient_a} \\
    =& \mathbb{E}_{p_{\alpha_a}(\textbf{z}^a)}\left[\nabla_{\alpha_a} U_{\alpha_a}(\textbf{z}^a)\right] -\mathbb{E}_{p_{\beta}(\textbf{z}^a|\textbf{I},\boldsymbol{\xi})}\left[\nabla_{\alpha_a} U_{\alpha_a}(\textbf{z}^a)\right], \nonumber
\end{align}
 the gradients for the EBM prior of object shape $\alpha_s$
\begin{align}
    &\mathbb{E}_{p_{\beta}(\textbf{z}^a, \textbf{z}^s|\textbf{I}, \boldsymbol{\xi})}\left[ \nabla_{\beta} \log p_{\alpha_s}(\textbf{z}^s)\right] \label{eq:prior_gradient_s}\\
    =& \mathbb{E}_{p_{\alpha_s}(\textbf{z}^s)}\left[\nabla_{\alpha_s} U_{\alpha_s}(\textbf{z}^s)\right] -\mathbb{E}_{p_{\beta}(\textbf{z}^s|\textbf{I},\boldsymbol{\xi})}\left[\nabla_{\alpha_s} U_{\alpha_s}(\textbf{z}^s)\right], \nonumber
\end{align}
as well as the gradients for the NeRF-based generator $\theta$
\begin{align}
    &\mathbb{E}_{p_{\beta}(\textbf{z}^a, \textbf{z}^s|\textbf{I}, \boldsymbol{\xi})}\left[\nabla_{\beta} \log p_{\theta}(\textbf{I}|\textbf{z}^a,\textbf{z}^s,\boldsymbol{\xi})\right] \label{eq:posterior_gradient}\\
    =&\mathbb{E}_{p_{\beta}(\textbf{z}^a, \textbf{z}^s|\textbf{I}, \boldsymbol{\xi})}\left[ \nabla_{\theta} G_{\theta}(\textbf{z}^a,\textbf{z}^s,\boldsymbol{\xi}) (\textbf{I}-G_{\theta}(\textbf{z}^a,\textbf{z}^s,\boldsymbol{\xi}))/{\sigma_{\epsilon}^2}\right]. \nonumber
\end{align}
Since the expectations in Eq.~(\ref{eq:prior_gradient_a}), Eq.~(\ref{eq:prior_gradient_s}), and  Eq.~(\ref{eq:posterior_gradient}) are analytically intractable, Langevin dynamics~\cite{neal2011mcmc}, which is a gradient-based MCMC sampling method, is employed to draw samples from the prior distributions (i.e., $p_{\alpha_a}(\textbf{z}^a)$ and $p_{\alpha_s}(\textbf{z}^s)$) and the posterior distribution (i.e., $p_{\beta}(\textbf{z}^a,\textbf{z}^s|\textbf{I},\boldsymbol{\xi})$), and then Monte Carlo averages are computed to estimate the expectation terms. As shown in Eq.~(\ref{eq:prior_gradient_a}) and Eq.~(\ref{eq:prior_gradient_s}), the update of the EBM prior model $\alpha_a$ (or $\alpha_s$) is based on the difference between $\textbf{z}^a$ (or $\textbf{z}^s$) sampled from the prior distribution $p_{\alpha_s}(\textbf{z}^a)$ (or $p_{\alpha_s}(\textbf{z}^s)$) and $\textbf{z}^a$ (or $\textbf{z}^s$) inferred from the posterior distribution  $p_{\beta}(\textbf{z}^a|\textbf{I},\boldsymbol{\xi})$ (or $p_{\beta}(\textbf{z}^s|\textbf{I},\boldsymbol{\xi})$).  According to Eq.(\ref{eq:posterior_gradient}), the update of the generator $\theta$ relies on $\textbf{z}^a$ and $\textbf{z}^s$ inferred from the posterior distribution $p_{\beta}(\textbf{z}^a, \textbf{z}^s|\textbf{I},\boldsymbol{\xi})$. To sample from the prior distributions $p_{\alpha_a}(\textbf{z}^a)$ and $p_{\alpha_s}(\textbf{z}^s)$ by Langevin dynamics, we update $\textbf{z}^a$ and $\textbf{z}^s$ by
\begin{align}
    \textbf{z}^a_{t+1} &= \textbf{z}^a_{t} + \delta \nabla_{\textbf{z}^a} \log p_{\alpha_a}(\textbf{z}^a_t) + \sqrt{2\delta} \textbf{e}^a_t, \label{eq:MCMC_prior_a}\\
        \textbf{z}^s_{t+1} &= \textbf{z}^s_{t} + \delta \nabla_{\textbf{z}^s} \log p_{\alpha_s}(\textbf{z}^s_t) + \sqrt{2\delta} \textbf{e}^s_t,
    \label{eq:MCMC_prior_s}
\end{align}
where $t$ indexes the time step, $\delta$ is the Langevin step size, and $\textbf{e}^a_t$ and $\textbf{e}^s_t$ are independent Gaussian noises that help the MCMC chains to escape from local modes during sampling. The gradients in Eq.~(\ref{eq:MCMC_prior_a}) and Eq.~(\ref{eq:MCMC_prior_s}) are given by
\begin{align}
    \nabla_{\textbf{z}^a} \log p_{\alpha_a}(\textbf{z}^a) &= -\nabla_{\textbf{z}^a} U_{\alpha_a}(\textbf{z}^a)-\textbf{z}^a/{\sigma^2}, \label{eq:MCMC_prior2_a}\\
         \nabla_{\textbf{z}^s} \log p_{\alpha_s}(\textbf{z}^s) &= -\nabla_{\textbf{z}^s} U_{\alpha_s}(\textbf{z}^s)-\textbf{z}^s/{\sigma^2}, 
    \label{eq:MCMC_prior2_s}
\end{align}
where $\nabla_{\textbf{z}^a} U_{\alpha_a}(\textbf{z}^a)$ and $\nabla_{\textbf{z}^s} U_{\alpha_s}(\textbf{z}^s)$ are efficiently computed by back-propagation.

For each observed $(\textbf{I}, \boldsymbol{\xi})$, we can sample from the posterior $p_{\beta}(\textbf{z}^a, \textbf{z}^s|\textbf{I},\boldsymbol{\xi})$ by alternately running Langevin dynamics: we fix $\textbf{z}^s$ and sample $\textbf{z}^a$ from $p_{\beta}(\textbf{z}^a|\textbf{z}^s,\textbf{I},\boldsymbol{\xi}) \propto p_{\beta}(\textbf{I},\textbf{z}^a|\textbf{z}^s,\boldsymbol{\xi})$, and then fix $\textbf{z}^a$ and sample $\textbf{z}^s$ from $p_{\beta}(\textbf{z}^s|\textbf{z}^a,\textbf{I},\boldsymbol{\xi}) \propto p_{\beta}(\textbf{I},\textbf{z}^s|\textbf{z}^a,\boldsymbol{\xi})$. The Langevin sampling step follows 
\begin{align}
    \textbf{z}^a_{t+1} &= \textbf{z}^a_{t} + \delta \nabla_{\textbf{z}^a} \log p_{\beta}(\textbf{I},\textbf{z}^a_t|\textbf{z}^s_t, \boldsymbol{\xi}) + \sqrt{2\delta} \textbf{e}^a_t, \label{eq:MCMC_posterior_a}\\
        \textbf{z}^s_{t+1} &= \textbf{z}^s_{t} + \delta \nabla_{\textbf{z}^s} \log p_{\beta}(\textbf{I},\textbf{z}^s_t|\textbf{z}^a_t, \boldsymbol{\xi}) + \sqrt{2\delta} \textbf{e}^s_t.
    \label{eq:MCMC_posterior_s}
\end{align}
The key steps in Eq.~(\ref{eq:MCMC_posterior_a}) and Eq.~(\ref{eq:MCMC_posterior_s}) are to compute the gradients of
\begin{align}
    \log & p_{\beta}(\textbf{I},\textbf{z}^a|\textbf{z}^s, \boldsymbol{\xi}) = \log  [p_{\alpha_a}(\textbf{z}^a)p_{\theta}(\textbf{I}|\textbf{z}^a,\textbf{z}^s,\boldsymbol{\xi})]=C_a\notag \\
    & -||\textbf{I}-G_{\theta}(\textbf{z}^a,\textbf{z}^s,\boldsymbol{\xi})||^2/2 \sigma_{\epsilon}^2-U_{\alpha_a}(\textbf{z}^a)-||\textbf{z}^a||^2/{2\sigma^2},\notag\\
    \log & p_{\beta}(\textbf{I},\textbf{z}^s|\textbf{z}^a, \boldsymbol{\xi}) =\log  [p_{\alpha_s}(\textbf{z}^s)p_{\theta}(\textbf{I}|\textbf{z}^s,\textbf{z}^a,\boldsymbol{\xi})] =C_s  \notag\\
    &-||\textbf{I}-G_{\theta}(\textbf{z}^a,\textbf{z}^s,\boldsymbol{\xi})||^2/2 \sigma_{\epsilon}^2-U_{\alpha_s}(\textbf{z}^s)-||\textbf{z}^s||^2/{2\sigma^2},\notag
\end{align}
where $C_a$ and $C_s$ are constants independent of $\textbf{z}^a$, $\textbf{z}^s$ and $\theta$. After sufficient alternating Langevin steps, the updated $\textbf{z}^a$ and $\textbf{z}^s$ follow the joint posterior $p_{\beta}(\textbf{z}^a, \textbf{z}^s|\textbf{I},\boldsymbol{\xi})$, and $\textbf{z}^a$ and $\textbf{z}^s$ follow $p_{\beta}(\textbf{z}^a|\textbf{I},\boldsymbol{\xi})$ and $p_{\beta}(\textbf{z}^s|\textbf{I},\boldsymbol{\xi})$, respectively.  

Let $\textbf{z}^{a-}_i$ and $\textbf{z}^{s-}_i$ be the samples drawn from the EBM priors by Langevin dynamics in Eqs.~(\ref{eq:MCMC_prior_a}) and ~(\ref{eq:MCMC_prior_s}). Let $\textbf{z}^{a+}_i$ and $\textbf{z}^{s+}_i$ be the inferred latent variables of the observation $(\textbf{I}_i,\boldsymbol{\xi}_i)$ by Langevin dynamics in Eqs.~(\ref{eq:MCMC_posterior_a}) and ~(\ref{eq:MCMC_posterior_s}). The gradients of the log-likelihood $L$ over $\alpha_a$, $\alpha_s$, and $\theta$ are estimated by
\begin{align}  
     &\nabla_{\alpha_a}{L}
     =\frac{1}{n}\sum_{i=1}^{n}\left[
     \nabla_{\alpha_a} U_{\alpha_a}(\textbf{z}^{a-}_i)\right]- \frac{1}{n}\sum_{i=1}^{n} \left[\nabla_{\alpha_a} U_{\alpha_a}(\textbf{z}^{a+}_{i})\right],  \notag\\
    &\nabla_{\alpha_s}{L}= \frac{1}{n}\sum_{i=1}^{n}\left[
     \nabla_{\alpha_s} U_{\alpha_s}(\textbf{z}^{s-}_i)\right]- \frac{1}{n}\sum_{i=1}^{n} \left[\nabla_{\alpha_s} U_{\alpha_s}(\textbf{z}^{s+}_{i})\right],  \notag\\
    &\nabla_{\theta}{L}=\frac{1}{n}\sum_{i=1}^{n}\left[ \nabla_{\theta} G_{\theta}(\textbf{z}^{a+}_i, \textbf{z}^{s+}_i,\boldsymbol{\xi}_i) \frac{\textbf{I}_i-G_{\theta}(\textbf{z}^{a+}_i, \textbf{z}^{s+}_i,\boldsymbol{\xi}_i)}{\sigma_{\epsilon}^2}\right].
    \notag
\end{align}
The learning algorithm of the NeRF-LEBM with MCMC inference can be summarized in Algorithm~\ref{alg:abp}.

\begin{algorithm}[t]
\caption{Learning NeRF-LEBM with MCMC inference}
\textbf{Input}: (1) Images and viewpoints $\{(\textbf{I}_i,\boldsymbol{\xi}_i)\}_{i=1}^n$; (2) Numbers of Langevin steps for priors and posterior $\{K^{-},K^{+}\}$; (3) Langevin step sizes for priors and posterior$\{\delta^{-},\delta^{+}\}$; 
(4) Learning rates for priors and generator $\{\eta_\alpha,\eta_\theta\}$.\\
\textbf{Output}: 
(1) $\theta$ for generator; (2) $(\alpha_a,\alpha_s)$ for EBM priors; (3) Latent variables $\{(\textbf{z}^a_i,\textbf{z}^s_i)\}_{i=1}^n$. 
\begin{algorithmic}[1]
\State Randomly initialize $\theta$, $\alpha_a$, $\alpha_s$, and $\{(\textbf{z}^a_i,\textbf{z}^s_i)\}_{i=1}^n$.
\Repeat
\State For each $(\textbf{I}_i,\boldsymbol{\xi}_i)$, sample the prior of object appearance $\textbf{z}_i^{a-} \sim p_{\alpha_a}(\textbf{z}^a)$ and the prior of object shape $\textbf{z}_i^{s-} \sim p_{\alpha_s}(\textbf{z}^s)$ using $K^{-}$ steps of Langevin dynamics with a step size $\delta^{-}$, which follows Eq.~(\ref{eq:MCMC_prior_a}) and Eq.~(\ref{eq:MCMC_prior_s}), respectively.
\State For each $(\textbf{I}_i,\boldsymbol{\xi}_i)$, run $K^+$ Langevin steps with a step size $\delta^{+}$, to alternatively sample $\textbf{z}^a_i$ from $p_{\beta}(\textbf{z}_i^a|\textbf{z}_i^s,\textbf{I}_i,\boldsymbol{\xi}_i)$, while fixing $\textbf{z}^s_i$; and sample $\textbf{z}_i^s$ from $p_{\beta}(\textbf{z}_i^s|\textbf{z}_i^a,\textbf{I}_i,\boldsymbol{\xi}_i)$, while fixing $\textbf{z}_i^a$. 
\State $\alpha_a \leftarrow \alpha_a+\eta_{\alpha} \nabla_{\alpha_a} {L}$. 
\State $\alpha_s \leftarrow \alpha_s+\eta_{\alpha}
\nabla_{\alpha_s}L$.
\State $\theta \leftarrow \theta+\eta_{\theta}\nabla_{\theta}L$.
\Until converge
\end{algorithmic} \label{alg:abp}

\end{algorithm}

\begin{algorithm}[t]
\caption{Variational Learning for NeRF-LEBM}
\textbf{Input}: (1) Images and viewpoints $\{(\textbf{I}_i,\boldsymbol{\xi}_i)\}_{i=1}^n$; (2) Number of Langevin steps $K^-$ for priors; (3) Langevin step size for priors $\delta^{-}$; (4) Learning rates $\{\eta_\alpha,\eta_{\omega}\}$.\\
\textbf{Output}: 
(1) $\theta$ for generator; (2) $(\alpha_a,\alpha_s)$ for EBM priors; (3) $\phi$ for inference net. 
\begin{algorithmic}[1]
\State Randomly initialize $\theta$, $\phi$, $\alpha_a$, and $\alpha_s$.
\Repeat
\State For each $(\textbf{I}_i,\boldsymbol{\xi}_i)$, sample the priors $\textbf{z}_i^{a-} \sim p_{\alpha_a}(\textbf{z}^a)$ and $\textbf{z}_i^{s-} \sim p_{\alpha_s}(\textbf{z}^s)$ using $K^{-}$ Langevin steps with a step size $\delta^{-}$, which follow Eq.~(\ref{eq:MCMC_prior_a}) and Eq.~(\ref{eq:MCMC_prior_s}) respectively.
\State For each $(\textbf{I}_i,\boldsymbol{\xi}_i)$, sample $\textbf{z}^a \sim q_{\phi_a}(\textbf{z}^a|\textbf{I}_i,\boldsymbol{\xi}_i)$ and $\textbf{z}_i^s \sim q_{\phi_s}(\textbf{z}^s|\textbf{I}_i,\boldsymbol{\xi}_i)$ using the inference network.
\State $\alpha_a \leftarrow \alpha_a+ \eta_{\alpha} \nabla_{\alpha_a} \text{ELBO}$ ($\nabla_{\alpha_a} \text{ELBO}$ is in Eq.~(\ref{eq:prior_vae_a})).
\State $\alpha_s \leftarrow \alpha_s+\eta_{\alpha} \nabla_{\alpha_s} \text{ELBO}$ ($\nabla_{\alpha_s} \text{ELBO}$ is in Eq.~(\ref{eq:prior_vae_s})).
\State $\omega \leftarrow \omega+\eta_{\omega} \nabla_{\omega}\text{ELBO}$ ($\nabla_{\omega}\text{ELBO}$ is in Eq.~(\ref{eq:posterior_vae}), where $\omega=(\phi_a, \theta)$).

\Until converge
\end{algorithmic} \label{alg:vae}

\end{algorithm}

\subsection{Learning with amortized inference}

Even though both prior and posterior sampling require Langevin dynamics. Prior sampling is more affordable than posterior sampling because the network structure of $U_{\alpha_a}$ or $U_{\alpha_s}$ is much smaller than that of the NeRF-based generator $G_{\theta}$ and the posterior sampling need to perform back-propagation on $G_{\theta}$, which is time-consuming. In this section, we propose to train the NeRF-LEBM by adopting amortized inference, in which the posterior distributions, $p_{\beta}(\textbf{z}^a|\textbf{I},\boldsymbol{\xi})$ and $p_{\beta}(\textbf{z}^s|\textbf{I},\boldsymbol{\xi})$, are approximated by separate bottom-up inference networks with reparameterization trick, $q_{\phi_a}(\textbf{z}^a|\textbf{I},\boldsymbol{\xi})=\mathcal{N}(\textbf{z}^a|u_{\phi_a}(\textbf{I}|\boldsymbol{\xi}),\sigma_{\phi_a}(\textbf{I}|\boldsymbol{\xi}))$ and $q_{\phi_s}(\textbf{z}^s|\textbf{I},\boldsymbol{\xi})=\mathcal{N}(\textbf{z}^s|u_{\phi_s}(\textbf{I}|\boldsymbol{\xi}),\sigma_{\phi_s}(\textbf{I}|\boldsymbol{\xi}))$, respectively. We denote $\phi=(\phi_a,\phi_s)$ for notation simplicity.
The log-likelihood $\log p_{\beta}(\textbf{I}|\boldsymbol{\xi})$ is lower
bounded by the evidence lower bound (ELBO), which is given by
\begin{align}
&\text{ELBO}(\textbf{I}|\boldsymbol{\xi}; \beta, \phi) \notag \\
 =& \log p_{\beta}(\textbf{I}|\boldsymbol{\xi}) - \mathbb{D}_{\text{KL}}(q_{\phi_a}(\textbf{z}^a|\textbf{I},\boldsymbol{\xi})||p_{\beta}(\textbf{z}^a|\textbf{I},\boldsymbol{\xi})) \notag\\
& -\mathbb{D}_{\text{KL}}(q_{\phi_s}(\textbf{z}^s|\textbf{I},\boldsymbol{\xi})||p_{\beta}(\textbf{z}^s|\textbf{I},\boldsymbol{\xi}))\\
 =& -\mathbb{D}_{\text{KL}}(q_{\phi_s}(\textbf{z}^s|\textbf{I},\boldsymbol{\xi})||p_{\alpha^s}(\textbf{z}^s)) \notag\\
 &-\mathbb{D}_{\text{KL}}(q_{\phi_a}(\textbf{z}^a|\textbf{I},\boldsymbol{\xi})||p_{\alpha^a}(\textbf{z}^a))\notag\\
& +\mathbb{E}_{q_{\phi_a}(\textbf{z}^a|\textbf{I},\boldsymbol{\xi})q_{\phi_s}(\textbf{z}^s|\textbf{I},\boldsymbol{\xi})}[\log p_{\theta}(\textbf{I}|\textbf{z}^a,\textbf{z}^s,\boldsymbol{\xi})]    ,
\end{align}
where $\mathbb{D}_{\text{KL}}$ denotes the Kullback-Leibler divergence. We assume $p_{\alpha^s}(\textbf{z}^s)=p_{\alpha^s}(\textbf{z}^s|\boldsymbol{\xi})$ and $p_{\alpha^a}(\textbf{z}^a)=p_{\alpha^a}(\textbf{z}^a|\boldsymbol{\xi})$. For the EBM prior models, the learning gradients to update $\alpha_a$ and $\alpha_s$ are given by
\begin{align} 
   &\nabla_{\alpha_a}\text{ELBO}(\textbf{I}|\boldsymbol{\xi}; \beta, \phi) \label{eq:prior_vae_a}\\ =& \mathbb{E}_{p_{\alpha_a}(\textbf{z}^a)}\left[\nabla_{\alpha_a} U_{\alpha_a}(\textbf{z}^a)\right]
   -\mathbb{E}_{q_{\phi_a}(\textbf{z}^a|\textbf{I},\boldsymbol{\xi})}\left[\nabla_{\alpha_a} U_{\alpha_a}(\textbf{z}^a)\right] , \notag\\
    &\nabla_{\alpha_s}\text{ELBO}(\textbf{I}|\boldsymbol{\xi}; \beta, \phi) \label{eq:prior_vae_s}\\=& \mathbb{E}_{p_{\alpha_s}(\textbf{z}^s)}\left[\nabla_{\alpha_s} U_{\alpha_s}(\textbf{z}^s)\right]-\mathbb{E}_{q_{\phi_s}(\textbf{z}^s|\textbf{I},\boldsymbol{\xi})}\left[\nabla_{\alpha_s} U_{\alpha_s}(\textbf{z}^s)\right] \notag.
\end{align}
Let $\omega = (\phi,\theta)$ be the parameters of the inference networks and generator. The learning gradients of these models are
\begin{align}
&\nabla_{\omega} \text{ELBO}(\textbf{I}|\boldsymbol{\xi}; \beta, \phi) \label{eq:posterior_vae}\\
=& \nabla_{\omega} \mathbb{E}_{q_{\phi_a}(\textbf{z}^a|\textbf{I},\boldsymbol{\xi})q_{\phi_s}(\textbf{z}^s|\textbf{I},\boldsymbol{\xi})}
[\log p_{\theta}(\textbf{I}|\textbf{z}^a,\textbf{z}^s,\boldsymbol{\xi})] \notag\\
-&\nabla_{\omega} \mathbb{D}_{\text{KL}}(q_{\phi_a}(\textbf{z}^a|\textbf{I},\boldsymbol{\xi})||p_0(\textbf{z}^a)) - \nabla_{\omega} \mathbb{E}_{q_{\phi_a}(\textbf{z}^a|\textbf{I},\boldsymbol{\xi})}
[U_{\alpha_a}(\textbf{z}^a)] \notag \\
-&\nabla_{\omega} \mathbb{D}_{\text{KL}}(q_{\phi_s}(\textbf{z}^s|\textbf{I},\boldsymbol{\xi})||p_0(\textbf{z}^s)) - \nabla_{\omega} \mathbb{E}_{q_{\phi_s}(\textbf{z}^s|\textbf{I},\boldsymbol{\xi})}
[U_{\alpha_s}(\textbf{z}^s)]\notag 
\end{align}
The first term on the right hand size of Eq.(\ref{eq:posterior_vae}) is the reconstruction by the bottom-up inference encoders and the top-down generator. The second and the fourth terms are KL divergences between the inference model and the Gaussian distribution. These three terms form the learning objective of the original VAE. The variational learning of the NeRF-LEBM is given in~Algorithm~\ref{alg:vae}.

\subsection{Learning without ground truth camera pose}
\label{sec:learn_no_pose}
Many real world datasets do not contain camera pose information, therefore fitting the models from those datasets by using Algorithms~\ref{alg:abp} or~\ref{alg:vae} is not appropriate. In this section, we study learning the NeRF-LFBM model from images without knowing the ground truth camera poses, and generalizing Algorithm~\ref{alg:vae} to this scenario. We treat the unknown camera pose as latent variables and seek to infer it together with the shape and appearance variables in the amortized learning framework. In our experiments, we assume the camera is located on a sphere and the object is put in the center of the sphere. Therefore, the camera pose $\boldsymbol{\xi}$ can be interpreted as the altitude angle ${\xi}_1$ and
azimuth angle ${\xi}_2$. However, different from the shape and the appearance,  the camera pose is directional and can be better explained through a spherical representation~\cite{mardia1975statistics}. For implementation, instead of directly representing each individual angle, we represent its Sine and Cosine values that directly construct the corresponding rotation matrix that is useful for subsequent computation. Thus, each rotation angle $\xi_i$ is a two-dimensional unit norm vector located on a unit sphere.

Following the hyperspherical VAE in~\cite{davidson2018hyperspherical}, we use the von Mises-Fisher (vMF) distribution to model the posterior distribution of ${\xi}$. vMF can be seen as the Gaussian distribution on a hypershere. To model a hypershpere of dimension $m$, it is parameterized by a mean direction $\boldsymbol{\mu} \in \mathbb{R}^m$ and a concentration parameter $\kappa \in \mathbb{R}_{\geq 0}$. The probability density of the vMF is defined as 
$p_{\text{vMF}}({\xi}| \boldsymbol{\mu}, \kappa) = \mathcal{C}_m(\kappa){\rm exp}(\kappa \boldsymbol{\mu}^T \boldsymbol{\xi})$,
where $\mathcal{C}_m(\kappa) = \frac{\kappa^{m/2 - 1}}{{(2\pi)}^{m/2} \mathcal{I}_{m/2-1}(\kappa)}$, with $\mathcal{I}_{l}$ denoting modified Bessel function of the first kind at order $l$. 
For each angel $\xi$ in $\boldsymbol{\xi}$, we design an inference model as  $q_{\phi_{\xi}}(\xi|\textbf{I})=p_{\text{vMF}}({\xi}| \boldsymbol{\mu}_{\phi_{\xi}}(\textbf{I}), \kappa_{\phi_{\xi}}(\textbf{I}))$, where $\boldsymbol{\mu}_{\phi_{\xi}}(\textbf{I})$ and $\kappa_{\phi_{\xi}}(\textbf{I})$ are bottom-up networks with parameters $\phi_{\xi}$ that maps $\textbf{I}$ to $\boldsymbol{\mu}$ and $\kappa$. We assume the prior of ${\xi}$ to be a uniform distribution on the unit sphere (denoted as $U(S^{m-1})$), which is the special case of vMF with $\kappa = 0$.  The key to use the amortized inference is to compute the KL divergence between the posterior and the prior, which can follow
\begin{align}
    &\mathbb{D}_{\text{KL}}(p_{\text{vMF}}(\boldsymbol{\mu}, \kappa) ||U(S^{m-1}))\nonumber\\
    =& \kappa \frac{\mathcal{I}_{m/2}(\kappa)}{\mathcal{I}_{m/2-1}(\kappa)} + {\rm log}\, \mathcal{C}_m(\kappa) - {\rm log} \, {(\frac{2(\pi^{m/2})}{\Gamma{m/2}})}^{-1}. 
\end{align}
Besides, to compute the ELBO, we need to draw samples from the inference model $q_{\phi_{\xi}}(\xi|\textbf{I})$, which amounts to sampling from the vMF distribution. We follow the sampling procedure in ~\cite{davidson2018hyperspherical} for this purpose in our implementation. 

\section{Experiments}
\subsection{Datasets}

To evaluate the proposed NeRF-LEBM framework and the learning algorithms, we conduct experiments on three datasets. The Carla dataset is rendered by~\cite{schwarz2020graf} using the Carla Driving Simulator~\cite{dosovitskiy2017carla}. It contains 10k cars of different shapes, colors and textures. Each car has one 2D image rendered from one random camera pose. Another dataset is the ShapeNet~\cite{chang2015shapenet} Car dataset, which contains 2.1k different cars for training and 700 cars for testing. We use the images rendered by~\cite{sitzmann2019scene} and follow its split to separate the training and testing sets. Each car in the training set has 250 views and we only use 50 views of them for training. Each car in the testing set has 251 views. 
each image is associated with its camera pose information. 

\subsection{Random image synthesis}\label{sec:synthesis}

 \begin{figure}[h!]
    \centering
    \begin{subfigure}[b]{.90\linewidth}
    \centering
    \includegraphics[width=0.7\textwidth]{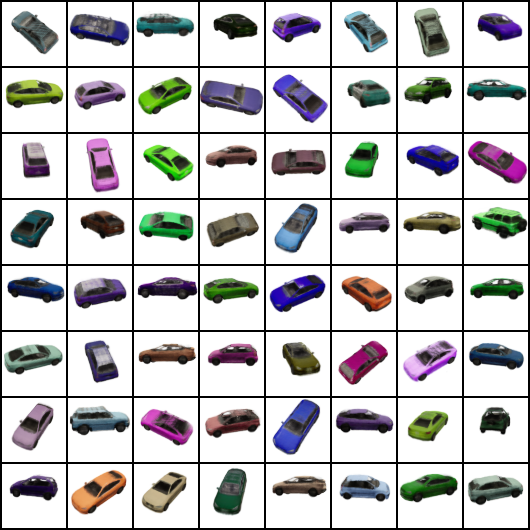}
    \caption{NeRF-LEBM with MCMC inference}
    \label{fig:Carla_gen_MCMC}
    \end{subfigure}
    \hfill
    \begin{subfigure}[b]{.90\linewidth}
    \centering
    \includegraphics[width=0.7\textwidth]{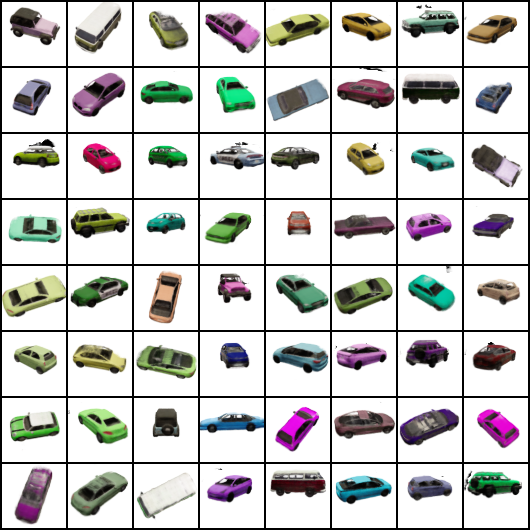}
    \caption{NeRF-LEBM with amortized inference}
    \label{fig:Carla_gen}
    \end{subfigure}
    \caption{Images generated by the NeRF-LEBM models trained on the Carla dataset, where the camera poses are given. (a) MCMC inference (b) amortized inference.}
    \label{fig:rand}
\end{figure}

\begin{figure*}[t!]
\centering
\begin{subfigure}[b]{.3\linewidth}
    \centering
    \includegraphics[width=0.9\textwidth]{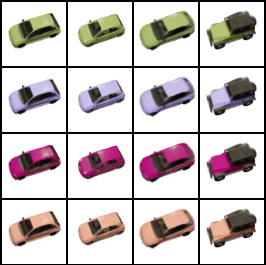}
    \caption{} \label{fig:Carla_interp}
\end{subfigure}
  \hspace{-0.0em}%
\begin{subfigure}[b]{.3\linewidth}
    \centering
    \includegraphics[width=0.9\textwidth]{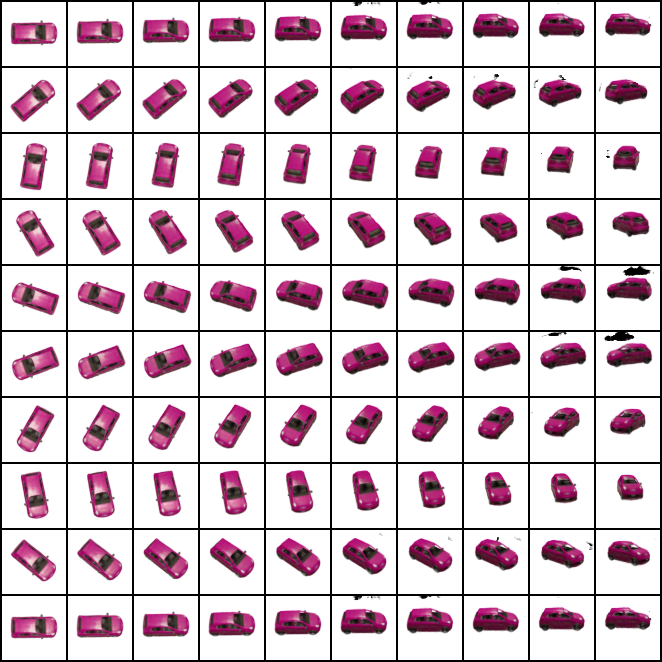}
    \caption{}\label{fig:Carla_different_view}
\end{subfigure}    
\hspace{-0.3em}%
\begin{subfigure}[b]{.38\linewidth}
    \centering
    \includegraphics[width=0.9\textwidth]{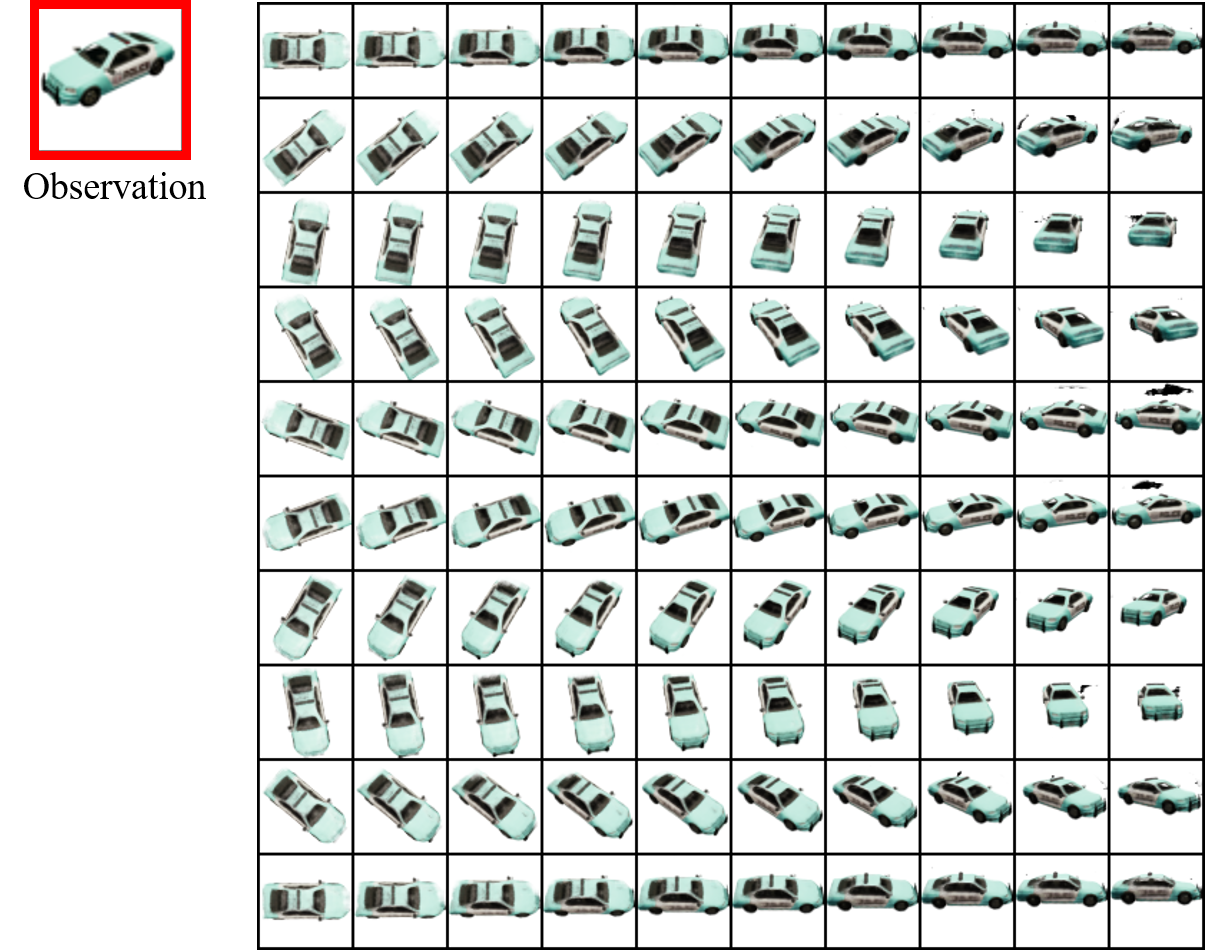}
    \caption{}\label{fig:Carla_novel_view}
\end{subfigure}

\caption{Disentangled representation. The generated images are obtained by the learned NeRF-LEBM using amortized inference on the Carla dataset.  (a) shows the influences of the shape vector $\textbf{z}^s$ and the appearance vector $\textbf{z}^a$ in image synthesis. The objects in each row share the same appearance vector $\textbf{z}^a$ and camera pose $\boldsymbol{\xi}$ but have different shape vectors $\textbf{z}^s$, while the objects in each column share the same shape vector $\textbf{z}^s$ and camera pose $\boldsymbol{\xi}$ but have different appearance vectors $\textbf{z}^a$. (b) demonstrates the effect of the camera pose variable $\boldsymbol{\xi}$ by varying it while fixing the shape and appearance vectors for a randomly sampled object. (c) shows an example of novel view synthesis for an observed 2D image.} 
\label{fig:Carla_dis}
\end{figure*}

We first evaluate the capability of image generation of the NeRF-LEBM on the Carla dataset, where the camera pose information is available. We try to answer whether the latent space EBMs can capture the underlying factors of objects in images and whether it is better than the Gaussian prior. We train our models on images of resolution $64 \times 64$ through both MCMC-based inference in Algorithm~\ref{alg:abp} and amortized inference in Algorithm~\ref{alg:vae}. Once a model is trained, we can generate new images by first randomly sampling $(\textbf{z}^a,\textbf{z}^s)$ from the learned EBM priors and a camera pose $\boldsymbol{\xi}$ from a uniform distribution, and then using the NeRF-based generator to map the sampled latent variables to the image space. The synthesized images by NeRF-LEBM using MCMC inference and amortized inference are displayed in Figures~\ref{fig:Carla_gen_MCMC} and ~\ref{fig:Carla_gen}, respectively. We can see the learned models can generate meaningful and highly diversified cars with different shapes, appearances and camera poses. To quantitatively evaluate the generative performance, we compare our NeRF-LEBMs with some baselines in terms of Fréchet inception distance (FID)~\cite{heusel2017gans} in Table~\ref{tab:Carla_fid}. The baselines include the NeRF-VAE~\cite{kosiorek2021nerf}, which is a NeRF-based generator using Gaussian prior and trained with variational learning, and the NeRF-Gaussian-MCMC, which is a NeRF-based generator using MCMC inference and Gaussian prior. To make a fair comparison, we implement the NeRF-VAE using the same NeRF-based generator and inference network as those in our NeRF-LEBM using amortized inference, except that the NeRF-VAE and the NeRF-Gaussian-MCMC only adopt Gaussian priors for latent variables. We compute FID using 10k samples.  Table~\ref{tab:Carla_fid} shows that NeRF-LEBMs perform very well in the sense that the learned models can generate realistic images. Especially, the NeRF-EBM trained with amortized inference obtains the best performance. The comparison between the NeRF-VAE and our NeRF-LEBM using amortized inference demonstrates the effectiveness of the EBM priors. The efficacy of the EBM priors is also validated by the comparison between the NeRF-Gaussian-MCMC and our NeRF-LEBM using MCMC inference.

\setlength{\tabcolsep}{4pt}
\begin{table}[b!]
\caption{Comparing the NeRF-LEBMs with likelihood-based baselines on the $64 \times 64$ Carla dataset for random image synthesis. The image qualities are evaluated via FID.}
\vspace{-0.05in}
\begin{center}
\begin{tabular}{ll}
    \hline\noalign{\smallskip}
    Likelihood-based Model & FID $\downarrow$\\
    \noalign{\smallskip}
    \hline
    \noalign{\smallskip}
    NeRF-Gaussian-MCMC & 54.13 \\
    NeRF-VAE~\cite{kosiorek2021nerf} & 38.15 \\
    NeRF-LEBM (MCMC inference) & 37.19 \\
    NeRF-LEBM (Amortized inference) & \textbf{20.84} \\
    \hline
\end{tabular}
\label{tab:Carla_fid}
\end{center}
\end{table}

\subsection{Disentangled representation}\label{sec:distentangle}

We investigate the ability of disentanglement of the NeRF-LEBM. We test the model using amortized inference trained in Section~\ref{sec:synthesis}. We generate images by varying one of the three latent vectors, i.e., $\textbf{z}^a$, $\textbf{z}^s$ and $\boldsymbol{\xi}$, while fixing the other two, and observe how the manipulated vector influences the generated images.  The generated images are shown in Figure~\ref{fig:Carla_dis}. In Figure \ref{fig:Carla_interp}, the objects in each row share the same appearance vector and camera pose but have different shape vectors, while the objects in each column share the same shape vector and camera pose but have different appearance vectors. From Figure~\ref{fig:Carla_interp}, we can see that the shape latent vectors do not encode any appearance information, such as color. The colors in the generated images only depend on the appearance latent vectors, and are not influenced by the shape latent vectors. Figure~\ref{fig:Carla_different_view} displays the synthesized images sharing the same appearance and shape vectors but having different camera poses. Figures~\ref{fig:Carla_interp} and ~\ref{fig:Carla_different_view} show that the learned model can successfully disentangle the appearance, shape and camera pose of an object because $\textbf{z}^a$, $\textbf{z}^s$ and $\boldsymbol{\xi}$ can, respectively, control the appearance, shape, and viewpoint of the generated images. We can also perform novel view synthesis of a seen 2D object by first inferring its appearance and shape vectors and then using different camera poses to generate different views of the object. Figure~\ref{fig:Carla_novel_view} shows one example. Please refer to Figure~\ref{fig:Carla_dis_MCMC} in Appendix for results obtained by a model trained with MCMC inference. 

\subsection{Inferring 3D structures of unseen 2D objects}\label{sec:unseen_obj}

\begin{figure}[h]
    \centering
    \includegraphics[width=.81\linewidth]{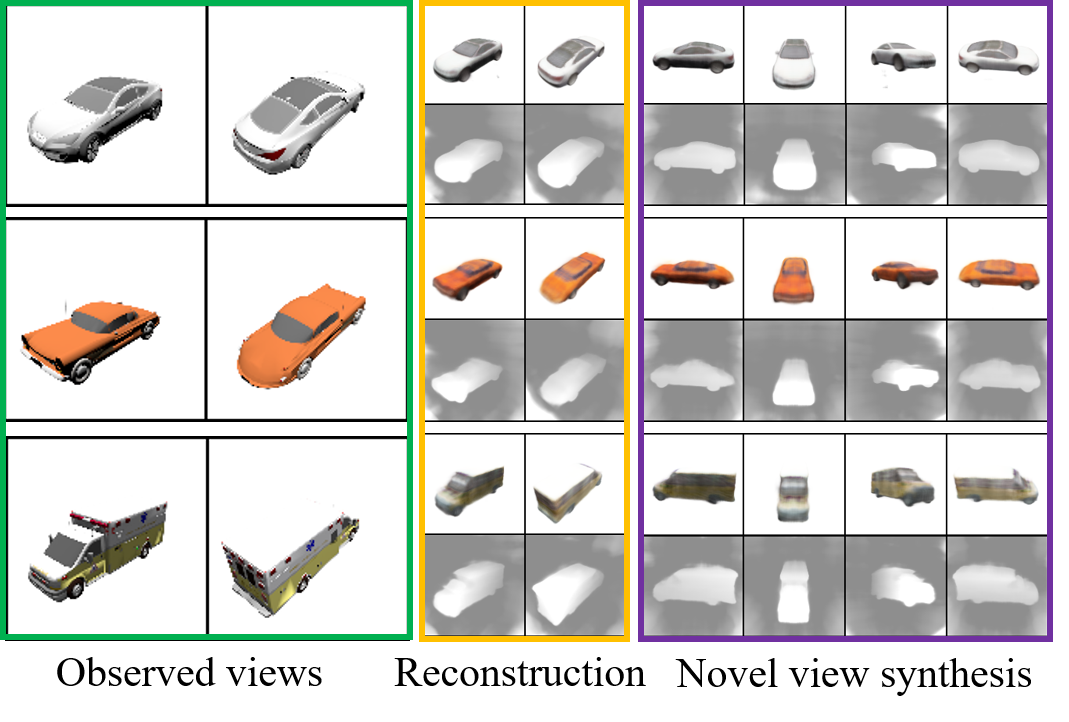}
    \caption{Two-shot novel view synthesis results on ($128 \times 128 $) the ShapeNet Car testing set. The left two columns displays two views of unseen cars in testing set. The middle two columns show the reconstruction results obtained by a model trained on training set. The right four columns shows novel view synthesis for the unseen cars.
    For each reconstructed or synthesized image, we show an RGB image at the first row and the inverse depth map at the second row.}
    \label{fig:SRN_2_shot}
\end{figure}

\begin{figure}[h]
    \centering
    \includegraphics[width=.81\linewidth]{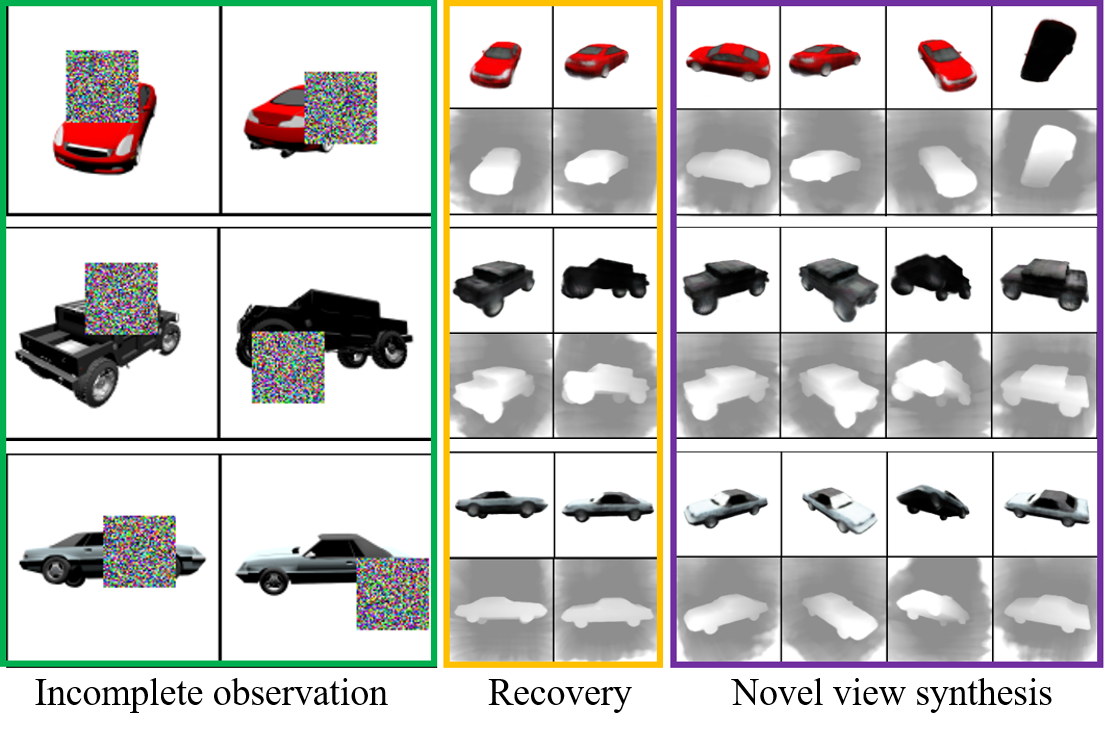}
    \caption{Learning from incomplete 2D observations. The left two columns show some examples of the incomplete observations in the training set. The middle two columns show the corresponding recovery results obtained by the learning algorithm. The right four columns are synthesis results for unobserved views of the same objects. For each recovery or synthesized result, we show an RGB 2D image at the first row and an inverse depth map at the second row.}
    \label{fig:SRN_mask}
\end{figure}

\begin{figure*}[h]
\centering

\begin{subfigure}{.31\linewidth}
    \centering
    \includegraphics[width=0.96\textwidth]{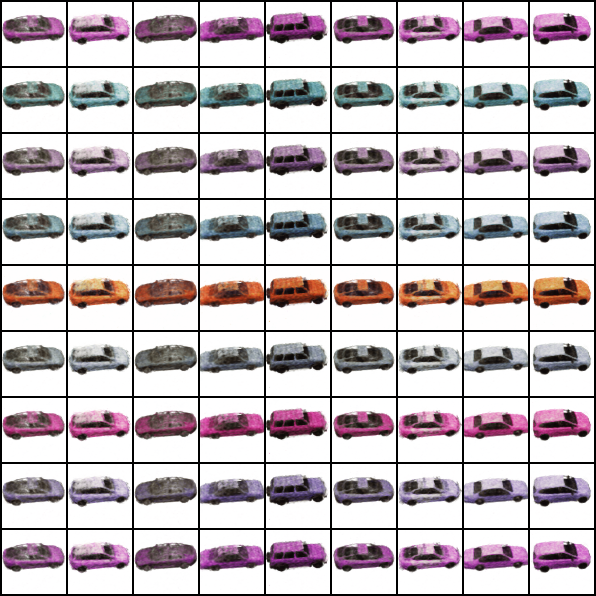}
    \caption{}\label{fig:unsup_interp_carla}
\end{subfigure}
\hspace{-0.2em}
\begin{subfigure}{.31\linewidth}
    \centering
    \includegraphics[width=0.96\textwidth]{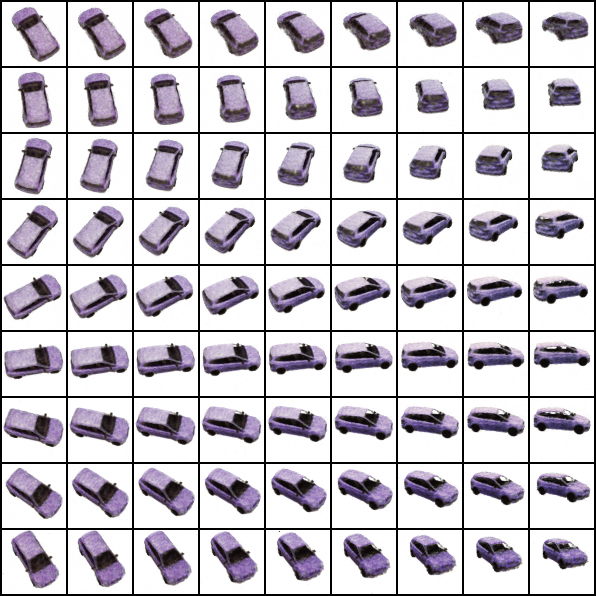}
    \caption{}
    \label{fig:unsup_pose_carla}
\end{subfigure}
\hspace{-0.2em}
\begin{subfigure}{.31\linewidth}
    \centering
    \includegraphics[width=0.96\textwidth]{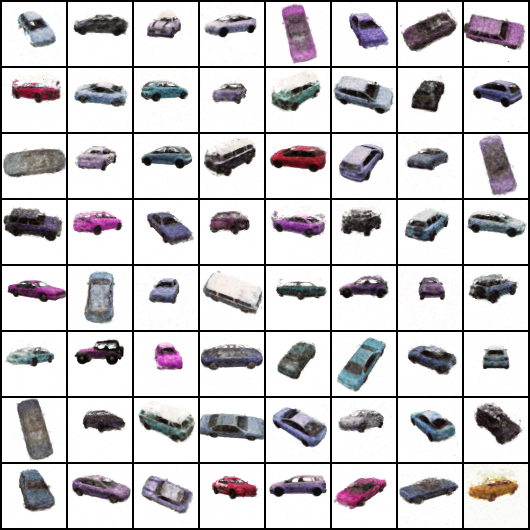}
    \caption{} \label{fig:unsup_syn_carla}
\end{subfigure}

\caption{Synthesis results on $64 \times 64$ Carla dataset without camera poses. (a) The objects in each row share the same appearance vector $\textbf{z}^a$ but have different shape vectors $\textbf{z}^s$ while the objects in each column share the same $\textbf{z}^s$ but have different $\textbf{z}^a$. They all share the same camera pose $\boldsymbol{\xi}$. (b) The Effect of changing the camera pose while fixing the shape and appearance vectors for a sampled object.  (c) Generated samples by randomly sampling $\textbf{z}^a,$ $\textbf{z}^s$ and camera pose $\boldsymbol{\xi}$.}
\label{fig:unsup_carla}
\end{figure*}

\begin{figure}[h]
    \centering
    \includegraphics[trim={0 260 260 0},clip, width=.3\textwidth]{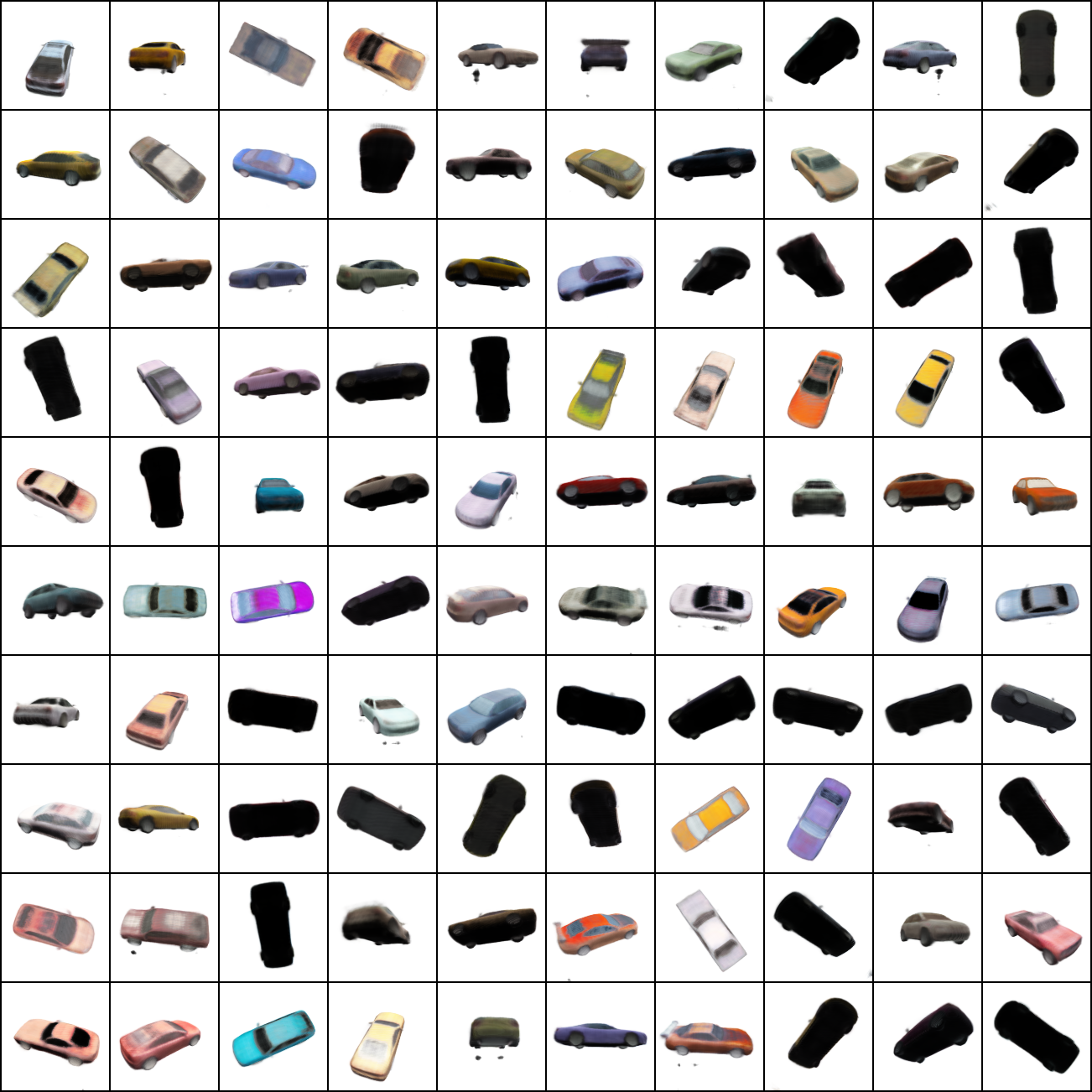}
    \caption{Synthesis by the NeRF-LEBM trained on the ShapeNet Car data, with persistent chain MCMC~inference.}
    \label{fig:fid_sample_deonly}
\end{figure}

Once a NeRF-LEBM model is trained, it is capable of inferring the 3D structure of a previously unseen object from only a few observations. Following~\cite{sitzmann2019scene}, we first train our model on the $128 \times 128$ resolution  ShapeNet Car training set and then apply the trained model to a two-shot novel view synthesis task, in which only two views of an unseen car in the holdout testing set are given to synthesize novel views of the same car. We use the NeRF-LEBM with MCMC inference in this task.To reduce the computational cost, we follow \cite{XieGZZW19} to adopt a persistent MCMC chain setting for the Langevin inference. Given a unseen object, we first infer its latent appearance and shape vectors via 600 steps of MCMC guided by the posterior distribution, and then we generate novel views of the same object with randomly sampled camera poses. The qualitative results of novel view synthesis are shown in Figure~\ref{fig:SRN_2_shot}. In  Table~\ref{tab:PSNR_car}, we compare our NeRF-LEBM with baselines, such as GQN~\cite{eslami2018neural} and the NeRF-VAE, in terms of PSNR. Our model has the best performance. We also demonstrate the generative ability of our model by showing the generated images 
in Figure~\ref{fig:fid_sample_deonly}. 
For a quantitative comparison, the FID for our NeRF-LEBM is 94.486 while the one obtained by the NeRF-VAE is~112.465.

\begin{table}[h]
\caption{Comparing the NeRF-LEBM with other baselines on two-shot novel view synthesis of unseen objects in terms of PSNR. Models are tested on the ShapeNet Car dataset.
}
\label{tab:PSNR_car}
\center
\begin{tabular}{ll}
\hline\noalign{\smallskip}
Generative Models &  PSNR $\uparrow$ \\
\noalign{\smallskip}
\hline
\noalign{\smallskip}
GQN~\cite{eslami2018neural} & 18.79 \\ 
NeRF-VAE~\cite{kosiorek2021nerf} & 18.37 \\ 
NeRF-LEBM (MCMC inference) & \textbf{20.28}\\   
\hline

\end{tabular}

\end{table}

\subsection{Learning from incomplete 2D observations}\label{sec:incomplete} 
To show the advantage of the MCMC-based inference, we test our model on a task where observations are incomplete or masked. To create a dataset for this task, we firstly randomly select 500 cars from the original ShapeNet Car dataset, and then for each of them, we use 50 views for training and 200 views for testing. For each training image, we randomly mask an area with Gaussian noise (see Figure~\ref{fig:SRN_mask}). To enable our model to learn from incomplete data, we only maximize the  data likelihood computed on the unmasked areas of the training data. This only leads to a minor modification in Algorithm~\ref{alg:abp} involving the computation of $||\textbf{I}-G_{\theta}(\textbf{z}^a,\textbf{z}^s,\boldsymbol{\xi})||^2$ in the likelihood term.  For a partially observed image, we compute it by summing over only the visible pixels. The latent variables can still be inferred by explaining the visible parts of the incomplete observations, and the model
can still be updated as before. In each iteration, we feed our model with two randomly selected observations of each object. 
Qualitative results in Figure~\ref{fig:SRN_mask} demonstrate that our algorithm can learn from incomplete images while recovering the missing pixels,  and the learned model can still perform novel view synthesis.  
We quantitatively compare our NeRF-LEBM with the NeRF-VAE in Table~\ref{tab:PSNR_FID_car_mask}. 
Our model beats the baseline using the same generator in the tasks of novel view synthesis and image~generation.

\begin{table}[h]
\caption{Comparison on tasks of novel view synthesis and image generation after learning the models from incomplete $128 \times 128$ 2D observations on a  masked ShapeNet Car dataset. PSNRs and FIDs are reported to measure the model performance on the two tasks, respectively.}
\label{tab:PSNR_FID_car_mask}\vspace{-0.1in}
\begin{center}
\begin{tabular}{lll}
\hline\noalign{\smallskip}
Model &  PSNR $\uparrow$ & FID $\downarrow$\\
\noalign{\smallskip}
\hline
\noalign{\smallskip}
NeRF-VAE & 21.26 & 128.27\\
NeRF-LEBM & \textbf{24.95} & \textbf{105.82}\\
\hline
\end{tabular}
\end{center}
\end{table}

\subsection{Learning with unknown camera poses}\label{sec:no_pose}

We study the scenario of training the NeRF-LEBM from 2D images without known camera poses. We assume the camera to locate on a sphere and the object is in the center of this sphere. Thus, we need to infer the altitude and azimuth angles for each observed image. As discussed in Section~\ref{sec:learn_no_pose}, we introduce a camera pose inference network to approximate the posterior of the latent camera pose of each observation, and train EBM priors, NeRF-based generator and inference networks simultaneously via amortized inference. In practice, we let the inference networks for camera pose $\boldsymbol{\xi}$, appearance $\textbf{z}^a$ and shape $\textbf{z}^s$ to share the lower layers and only differ in their prediction heads. We carried out an experiment on the Carla dataset with a resolution of $64 \times 64$. Unlike in Section \ref{sec:synthesis}, here we only use the rendered images and do not use the ground truth camera pose associated with each image. The results on are shown in Figure~\ref{fig:unsup_carla}. From Figure~\ref{fig:unsup_interp_carla}, we can see the learned model can disentangle shape and appearance factors. Figure~\ref{fig:unsup_pose_carla} shows that the learned model can factor out the camera pose from the data in an unsupervised manner.    Figure~\ref{fig:unsup_syn_carla} shows some random synthesized examples generated from the model by randomly sampling $\textbf{z}^a$, $\textbf{z}^s$ and $\boldsymbol{\xi}$. Although the camera pose can take a valid value from $[0, 2\pi)$, the altitude and azimuth angles of the training examples  in the Carla dataset might only lie in limited ranges. Thus, after training, we estimate a camera pose distribution using 10,000 training examples and when we generate images, we sample camera poses according to this distribution. Please refer to Figure~\ref{fig:unsup_celeba} in Appendix for more results on the CelebA dataset. These results verify that our model can learn meaningful 3D-aware 2D image generator without known camera poses.

\section{Conclusion and limitations}
This paper pushes forward the progress of development of  likelihood-based generative radiance field models for disentangled representation by proposing the NeRF-LEBM framework, in which we build informative and trainable energy-based priors for latent variables of object appearance and object shape on top of a NeRF-based top-down 2D image generator, and presenting two maximum likelihood learning algorithms, one with MCMC-based inference and the other with amortized inference. We formulate the NeRF-LEBM framework in two scenarios. One is training with known camera poses and the other is with unknown camera poses. Through several tasks we show that NeRF-LEBM can generate meaningful samples, infer 3D structure from 2D observations, learn to infer from incomplete 2D observation, and even learn from images with unknown camera poses. However, our current model also have some limitations. First, despite its accuracy and memory efficiency, the MCMC-based inference is indeed time-consuming. Thus we have to resort to amortized inference for efficient computation. 
Secondly, as for training from images with unknown camera poses, we find that it is still changeling for our model to obtain comparable generation performance with GAN-based models that can get around the inference step. Also, the training with unknown camera poses using amortized inference is still unstable and difficult.  We will continue to improve our model in the future~work.

\newpage 
\subsubsection*{Acknowledgements}
The authors sincerely thank Dr. Ying Nian Wu at statistics department of the University of California, Los Angeles (UCLA) for the helpful discussion on the topic of von Mises-Fisher distribution in directional statistics. The authors would also
like to thank the anonymous reviewers for providing constructive comments and suggestions to improve the work. Our work is also supported by XSEDE grant CIS210052, which provides GPU computing resources. 

\bibliographystyle{plain}
\bibliography{refs_scholar}

\appendix
\onecolumn
\section{Training details}
In this section, we describe the network structure designs and hyper-parameter settings in our experiments.
\subsection{Network structure}
\subsubsection{Conditional NeRF-based generator} 
\begin{figure*}[b]
\centering
\begin{subfigure}[b]{.35\linewidth}
    \centering
    \includegraphics[width=0.99\textwidth]{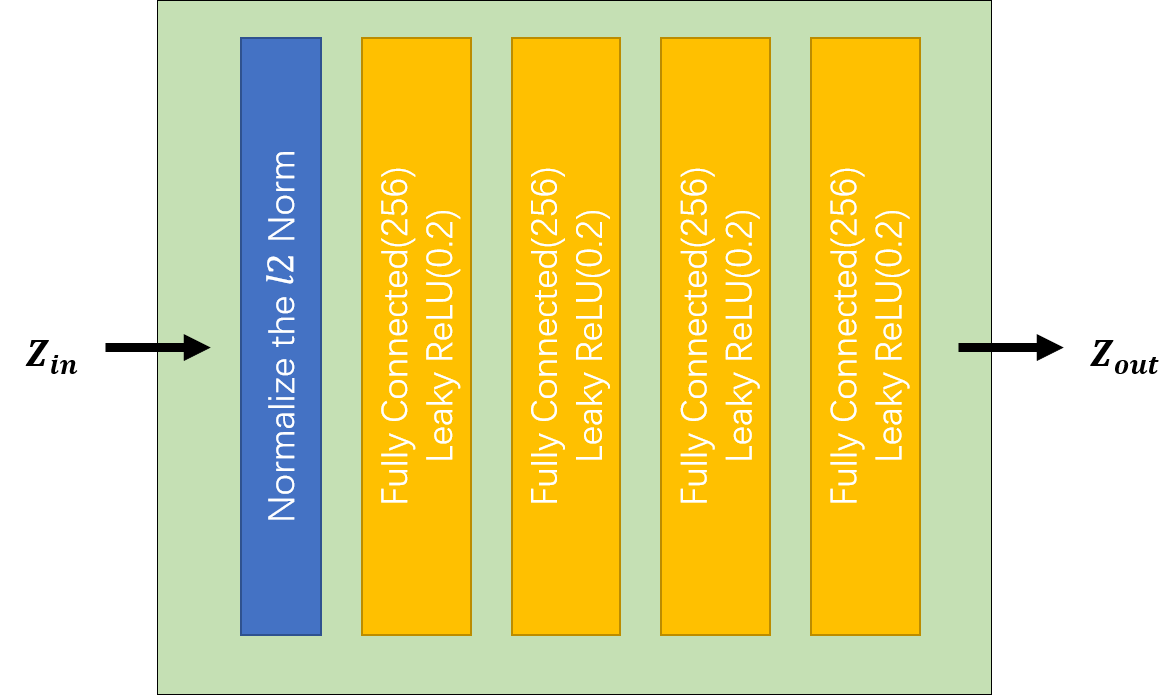}
    \caption{Mapping network} \label{fig:latent_encoder}
\end{subfigure}
  \hspace{-1.5em}%
\begin{subfigure}[b]{.65\linewidth}
    \centering
    \includegraphics[width=0.95\textwidth]{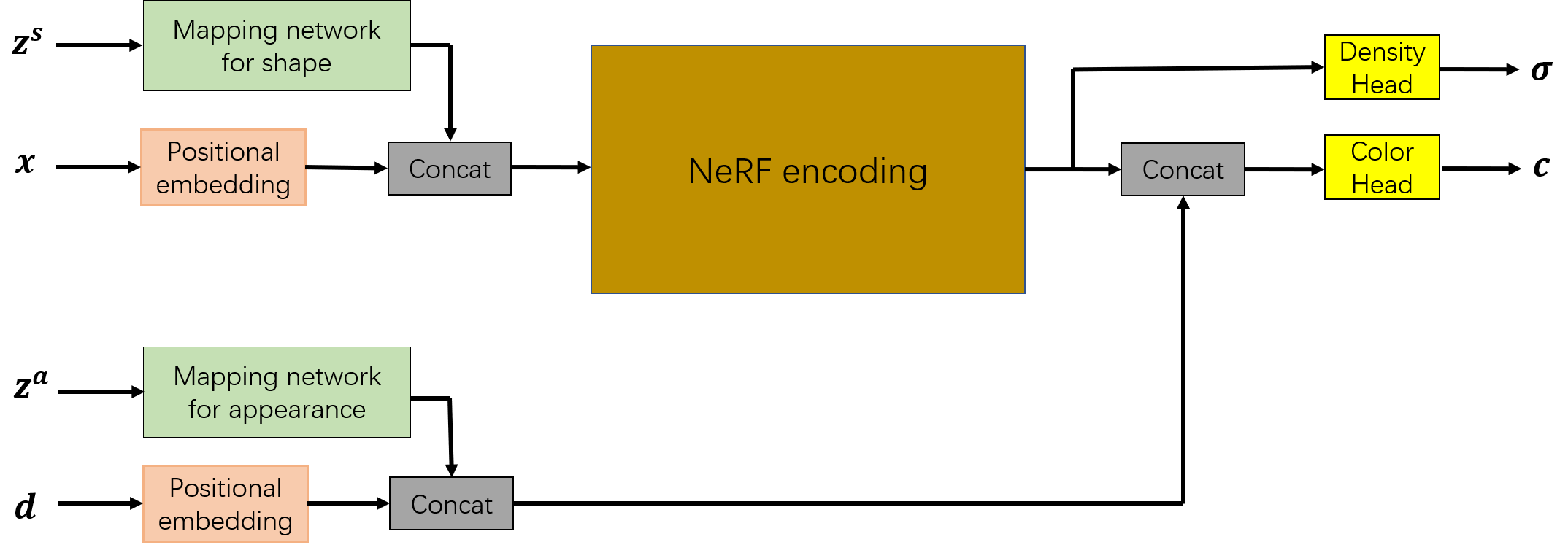}
    \caption{Conditional NeRF-based generator.}\label{fig:condi_nerf_gen}
\end{subfigure}    

\caption{Model structure for the NeRF-LEBM generator with known camera pose.} 
\label{fig:NeRF structure}
\end{figure*}

\begin{figure*}[ht]
\centering
\begin{subfigure}[t]{.3\linewidth}
    \centering
    \includegraphics[width=0.95\textwidth]{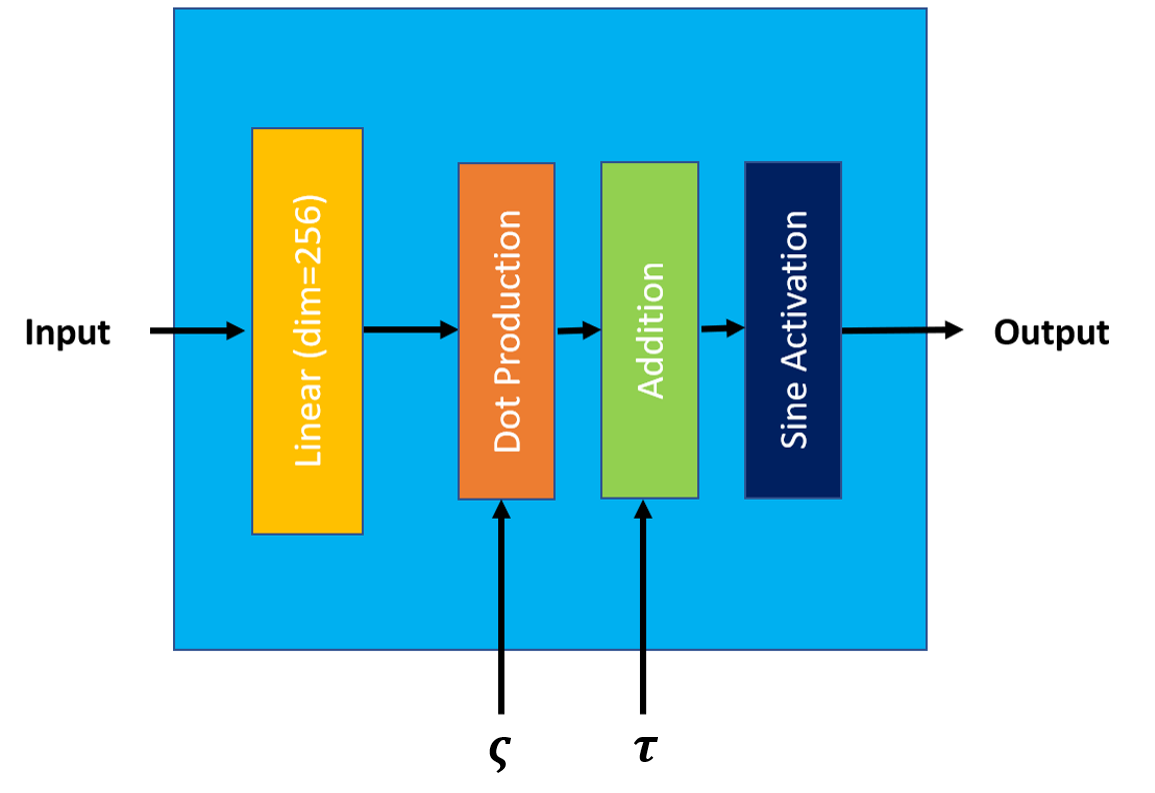}
    \caption{The FiLM layer.} \label{fig:latent_encoder}
\end{subfigure}
  \hspace{-0.2em}%
\begin{subfigure}[t]{.3\linewidth}
    \centering
    \includegraphics[width=0.95\textwidth]{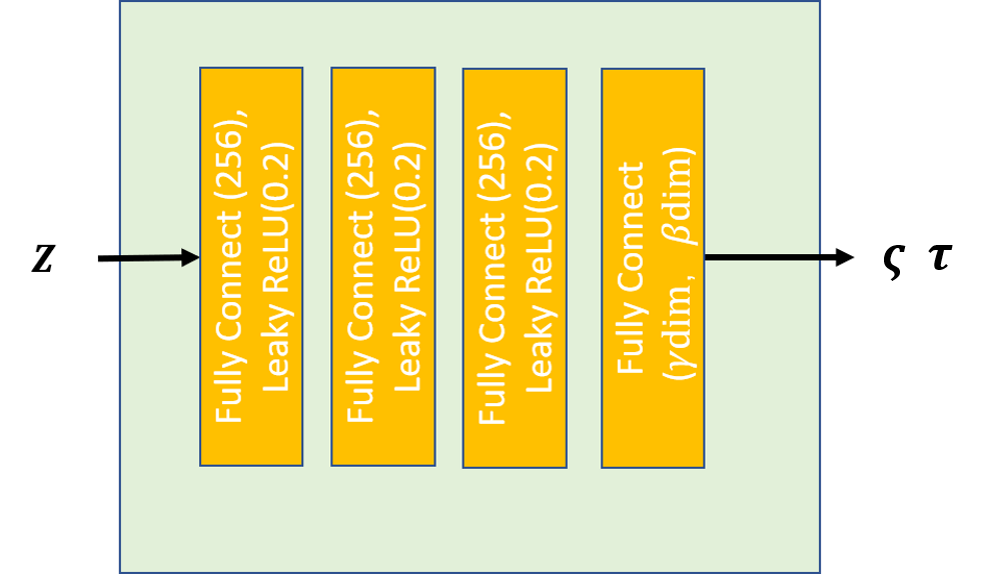}
    \caption{Mapping network.}\label{fig:condi_nerf_gen}
\end{subfigure}

\begin{subfigure}[b]{.8\linewidth}
    \centering
    \includegraphics[width=0.93\textwidth]{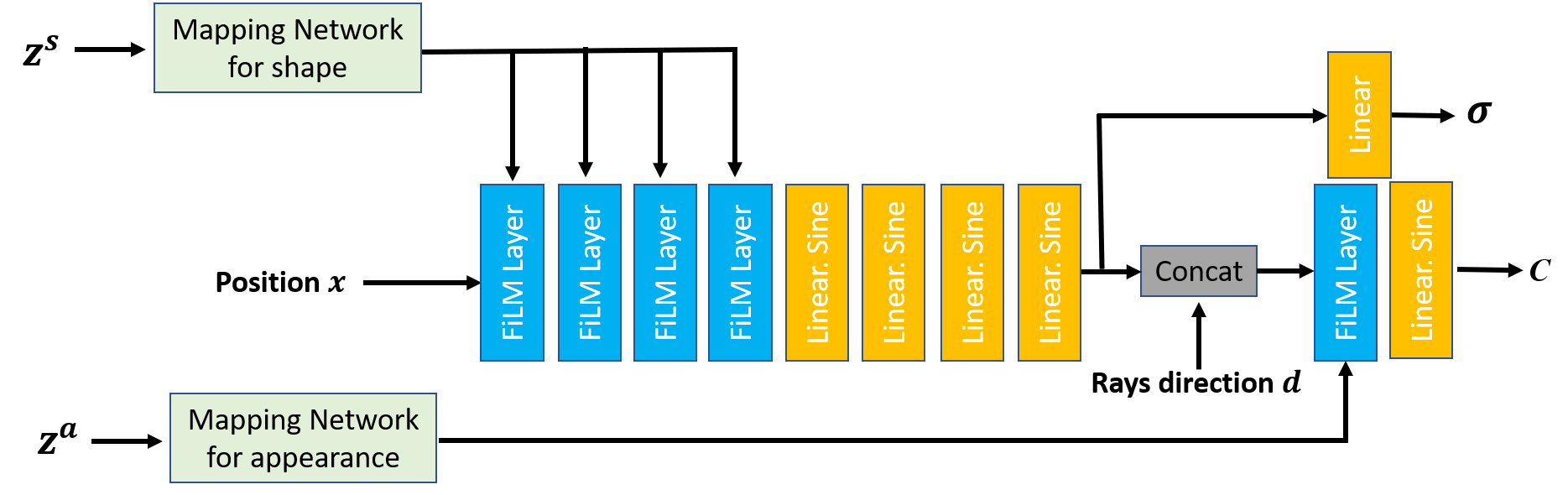}
    \caption{Conditional NeRF-based generator structure.}\label{fig:condi_nerf_gen}
\end{subfigure}

\caption{Model structure for the NeRF-LEBM generator with unknown camera pose.} 
\label{fig:NeRF structure no pose}
\end{figure*}

For the experiments where camera poses of objects are given, the structure of the NeRF-based generator is shown in Figure~\ref{fig:NeRF structure}. For this structure, we mainly follow the design in~\cite{schwarz2020graf}. However, for each of the latent vectors $\textbf{z}^s$ and $\textbf{z}^a$, we add a mapping network to transform it before concatenating it with the positional embedding. This mapping network is composed of a normalizing layer and 4 linear layers, each of which is followed by a Leaky ReLU activation function. The concatenated features then enter the NeRF encoding module, which is an 8-layer MLP. The number of dimensions of each hidden layer or the output layer is 256. The output of the NeRF encoding module is then used to predict the color $\textbf{c}$ and density $\sigma$ information. For the experiments about learning without camera pose information on the CelebA dataset, inspired by~\cite{chan2021pi}, we design a network structure with the FiLM-conditioned layers~\cite{perez2018film,dumoulin2018feature}. To be more specific, we input the transformed $\textbf{z}^s$ to the first 4 layers of the generator and input the transformed $\textbf{z}^a$ to the color head using the FiLM layer. The detailed architecture is shown in Figure~\ref{fig:NeRF structure no pose}. 

\subsubsection{Latent space EBM}

We use two EBM structures. One (small version) contains 2 layers of linear transformations, each of which followed by a swish activation, while the other (large version) contains 4 layers of linear transformations with swish activation layers. The network structure is shown in Table~\ref{tab:LEBM}. The choice of the EBM prior model for each experiment is shown in Table~\ref{tab:EBM_each_exp}.

\subsubsection{Inference model} 
In the experiments using amortized inference, we need an inference model to infer the latent vectors (i.e., appearance, shape, and camera pose). Inspired by~\cite{brahmbhatt2018geometry}, we use the ResNet34~\cite{he2016deep} structure as the feature extractor and build separate inference heads on the top of it for different latent vectors. Each inference head is composed of a couple of MLP layers. We let the inference models for $\textbf{z}^a$, $\textbf{z}^s$ and camera pose to share the same feature extractor and only differ in their inference heads. When we update the parameters of the inference model, we use a learning rate $3 \times 10^{-5}$ for the feature extractor and the pose head while using a smaller learning rate $5 \times 10^{-6}$ for the inference heads of $\textbf{z}^a$ and $\textbf{z}^s$.

\subsection{Training hyperparameters}
During our training, we use the Adam\cite{kingma2015adam} optimizer and we set $\beta_{{\text{Adam}}} = (0.9, 0.999)$. We set the standard deviation of the residual in Eq. (\ref{eq:generator}) $ \sigma_\epsilon = 0.05$. We disable the noise term in the Langevin inference for a better performance in the MCMC inference case. For the persistent chain MCMC inference setting, we use the Adam instead of a noise-disable Langevin dynamics to infer the latent vectors. As to the MCMC sampling for the EBM priors, we assign a weight as a hyperparameter to the noise term in the Langevin dynamics for adjusting its magnitude in our practise. Please check Table~\ref{tab:Hyper-param} for hyperparameter settings, including learning rate, batch size, etc, used in other experiments.

\begin{table}[h]
\centering
\caption{Model structures of the latent EBMs with hidden dimension $C$.}
\begin{subtable}{0.25\textwidth}
\centering
\caption{EBM (small)}
\begin{tabular}{c}
\hline
Linear $C$, swish\\
Linear $C$, swish\\
Linear 1 \\
\hline
\end{tabular}
\label{tab:2_layer_EBM}
\end{subtable}
\hspace{-2em}
\begin{subtable}{0.25\textwidth}
\caption{EBM (large)}
\centering
\begin{tabular}{c}
\hline
Linear $C$, swish\\
Linear $C$, swish\\
Linear $C$, swish\\
Linear $C$, swish\\
Linear 1 \\
\hline
\end{tabular}

\label{tab:4_layer_EBM}
\end{subtable}

\label{tab:LEBM}
\end{table}

\begin{table*}[h!]
\centering
\caption{Choice of the EBM prior in each experiment.}
\begin{tabular}{ccccccc}
\hline\noalign{\smallskip}
\multirow{2}{*}{Experiment} & \multicolumn{3}{c}{EBM prior for shape} & \multicolumn{3}{c}{EBM prior for appearance} \\
& type & \makecell{vector \\ dimension} & \makecell{hidden \\ dimension} & type & \makecell{vector \\ dimension} & \makecell{hidden \\ dimension}\\
\noalign{\smallskip}
\hline
\noalign{\smallskip}
\makecell{Carla / MCMC inference / with poses} & \makecell{ large} & 128 & 256 & \makecell{large} & 128 & 256\\
\makecell{Carla / amortized inference / with poses} & \makecell{small} & 128 & 256 &  \makecell{small} & 128 & 128\\
\makecell{ShapeNet Car / MCMC inference / with poses} & \makecell{large} & 128 & 256 & \makecell{large} & 128 & 256 \\ 
\makecell{Carla / amortized inference / without poses} & \makecell{small} & 128 & 128 & \makecell{small} & 128 & 64 \\ 
\hline
\end{tabular}

\label{tab:EBM_each_exp}
\end{table*}

\begin{table*}[h]
\caption{Hyperparameter settings in different experiments.}
\centering
\begin{tabular}{cccccccccccc}
\hline\noalign{\smallskip}
 Experiements & \makecell{batch \\ size} & \makecell{Number \\ of views} & $\eta_{\alpha}$ & $\eta_{\theta}$ & $\eta_{\phi}$ & $q_0$ & $\delta^{+}$ & $K^{+}$ & $\delta^{-}$ & $K^{-}$ & \makecell{MCMC noise \\ weight (EBM)} \\
\noalign{\smallskip}
\hline
\noalign{\smallskip}
 \makecell{Carla / MCMC \\  inference / with poses} & 8 & 1 & 2 e-5 & 1e-4 & - & Uniform & 0.1 & 60 & 0.5 & 60 & 0.02 \\
 \makecell{Carla / amortized \\ inference / with poses } & 8 & 1 & 7e-6 & 1e-4 & 1e-4 & Normal & - & - & 0.5 & 60 & 1.0 \\
 \makecell{ShapeNet Car / persistent \\ MCMC inference / with poses} & 12 & 2 & 2e-5 & 1e-4 & -& Normal & 1e-4  & 1 & 0.5 & 40 & 0.0\\
 \makecell{Carla / amortized \\ inference / without poses} & 16 & 1 & 7e-6 & 3e-5 & 3e-5 & Normal & - & - & 0.5 & 60 & 0.0 \\
\hline
\end{tabular}

\label{tab:Hyper-param}
\end{table*}

\section{Complexity analysis}
We show a comparison of different models in terms of model size in Table~\ref{tab:model_size}. By comparing our NeRF-LEBM using amortized inference with the NeRF-VAE which uses a Gaussian prior, we can find that  introducing a latent EBM as a prior only slightly increase the number of parameters. If we rely on the MCMC of inference, then we can save a lot of parameters. In Table~\ref{tab:syn_time}, we compare the computational time of randomly sampling 100 images with a resolution of $128 \times 128$. The time is computed when the algorithm is run on a single NVIDIA RTX A6000 GPU. Comparing with the Gaussian prior, the EBM prior (with a 40-step MCMC sampling) barely affects the sampling time. That is because the latent EBM is much less computationally-intensive than the NeRF-based generator and we can sample 100 random variables in a batch altogether. We also compare the computational time of the amortized inference and MCMC inference in the few-shot inference of an unseen object in Table~\ref{tab:infer_time}, from which we can see that using an inference model can do a much quicker inference but the reconstruction results may be worse. On the other hand, MCMC inference may take some time but the results can be~better.

\begin{table}[h!]
\centering
\caption{Complexity analysis}
\begin{subtable}[t]{0.39\textwidth}
\centering
\caption{Model size}

\begin{tabular}{lc}
\toprule
Model & \# parameters \\
\midrule
NeRF-VAE & 24.06 M \\ 
NeRF-LEBM (amortized) & 24.52 M \\
NeRF-LEBM (MCMC) & 2.21 M\\
\bottomrule
\end{tabular}
\label{tab:model_size}
\end{subtable}
\hspace{-1em}
\begin{subtable}[t]{0.30\textwidth}
\centering
\caption{Synthesis time}

\begin{tabular}{lc}
\toprule
Model & Time \\
\midrule
EBM prior & 40.97s \\
Gaussian prior & 40.73s \\
\bottomrule
\end{tabular}
\label{tab:syn_time}
\end{subtable}
\hspace{-1em}
\begin{subtable}[t]{0.31\textwidth}
\centering
\caption{Inference time}
\begin{tabular}{lc}
\toprule
Model & Time \\
\midrule
Amortized inference & 0.1s \\
MCMC inference & 168s \\
\bottomrule
\end{tabular}
\label{tab:infer_time}
\end{subtable}
\label{tab:Complexity_analysis}
\end{table}

\section{More generation results}
\subsection{More results for the experiment in  Section~\ref{sec:synthesis} Random image synthesis}

\begin{figure}[h]
    \centering
    \includegraphics[width=.35\textwidth]{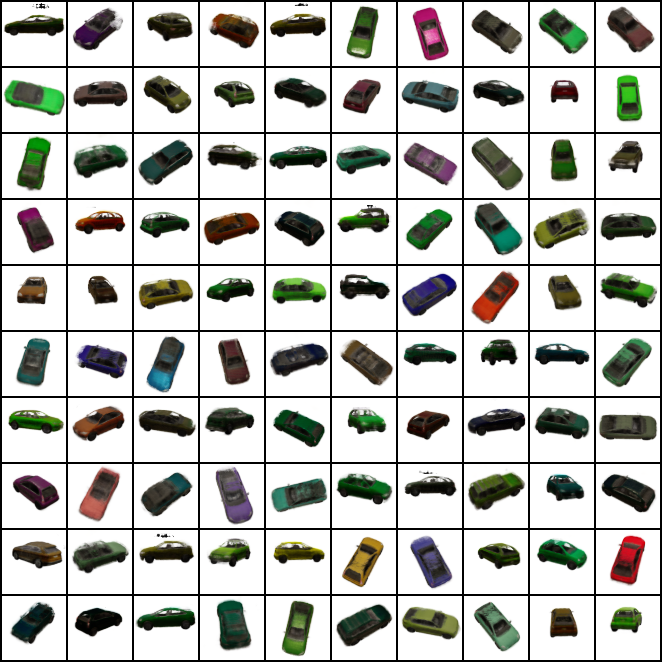}
    \caption{Images generated by the NeRF-Gaussian-MCMC baseline}
    \label{fig:carla_gen_MCMC_gaussian}
\end{figure}

We show the generated examples from the NeRF-Gaussian-MCMC in Figure~\ref{fig:carla_gen_MCMC_gaussian}. Comparing the results in Figure~\ref{fig:carla_gen_MCMC_gaussian} with those in Figure~\ref{fig:rand}, we can see that examples obtained from the simple Gaussian prior have less diversity than those from the EBM prior. This is consistent with the numerical evaluation shown Table~\ref{tab:Carla_fid} and demonstrates the advantage of using latent EBMs as informative prior distributions for modeling latent variables.

\subsection{More results for the experiment in Section~\ref{sec:distentangle} Disentangled representation}
 
\begin{figure}[h!]
    \centering
    \begin{subfigure}{.35\textwidth}
        \centering
        \includegraphics[width=.92\textwidth]{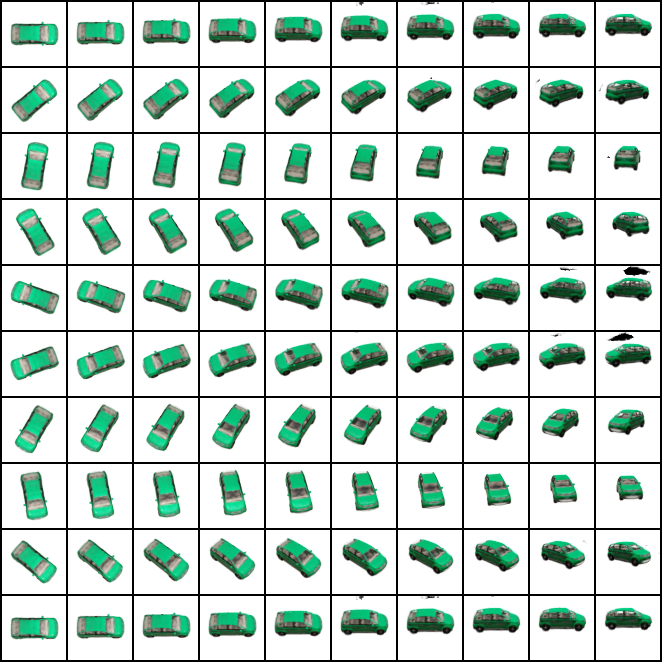}
    \end{subfigure}
    \begin{subfigure}{.35\textwidth}
        \centering
        \includegraphics[width=.92\textwidth]{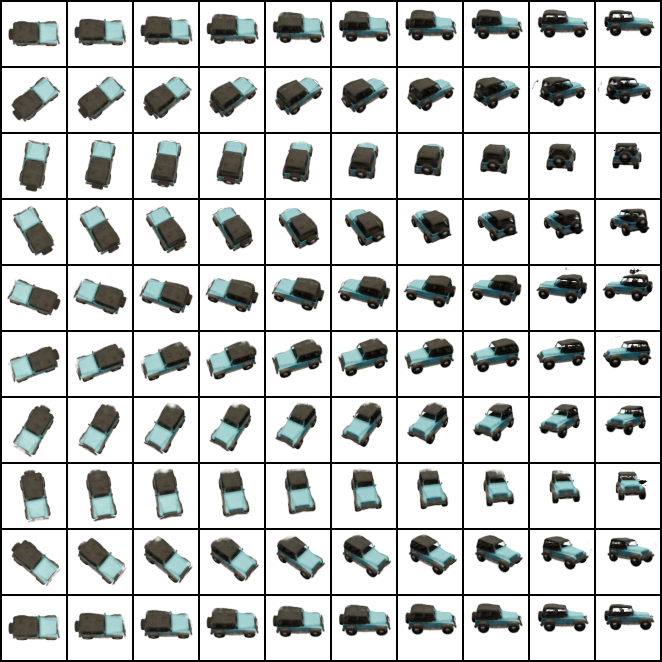}
    \end{subfigure}
    \caption{More generated examples by the NeRF-LEBM using amortized inference on the Carla dataset.}
    \label{fig:carla_gen}
\end{figure}

\begin{figure*}[h!]
 
\centering

\begin{subfigure}[b]{.3\linewidth}
    \centering
    \includegraphics[width=0.86\textwidth]{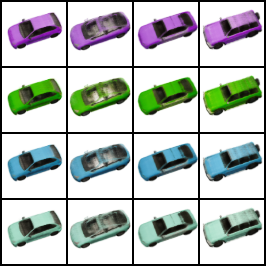}
    \caption{} \label{fig:Carla_interp_MCMC}
\end{subfigure}
  \hspace{-0.0em}%
\begin{subfigure}[b]{.3\linewidth}
    \centering
    \includegraphics[width=0.86\textwidth]{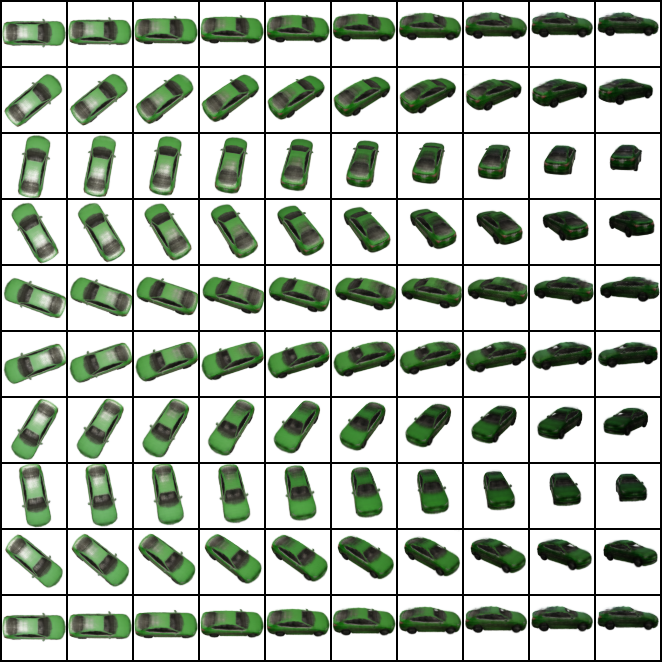}
    \caption{}\label{fig:Carla_different_view_MCMC}
\end{subfigure}    
\hspace{-0.3em}%
\begin{subfigure}[b]{.38\linewidth}
    \centering
    \includegraphics[width=0.9\textwidth]{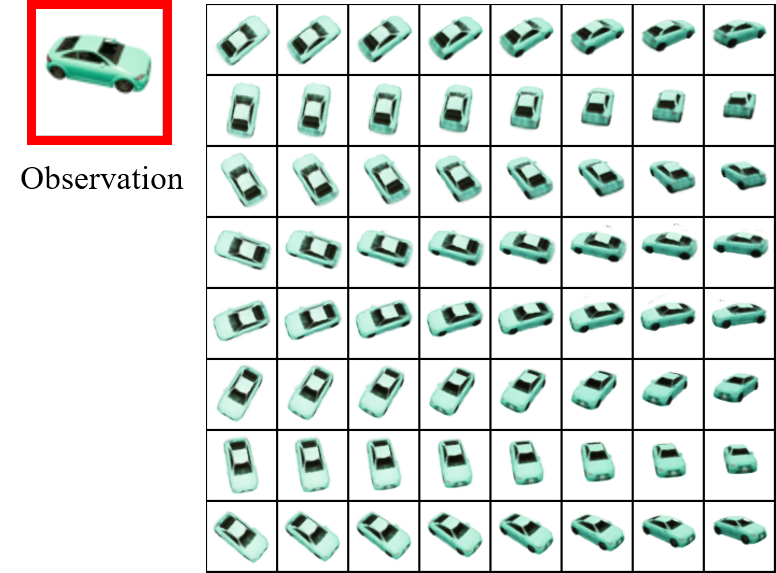}
    \caption{}\label{fig:Carla_novel_view_MCMC}
\end{subfigure}

\caption{Disentangled representation. The generated images are obtained by the learned NeRF-LEBM using MCMC inference on the Carla dataset.  (a) shows the influences of the shape vector $\textbf{z}^s$ and the appearance vector $\textbf{z}^a$ in image synthesis. The objects in each row share the same appearance vector $\textbf{z}^a$ and camera pose $\boldsymbol{\xi}$ but have different shape vectors $\textbf{z}^s$, while the objects in each column share the same shape vector $\textbf{z}^s$ and camera pose $\boldsymbol{\xi}$ but have different appearance vectors $\textbf{z}^a$. (b) demonstrates the effect of the camera pose variable $\boldsymbol{\xi}$ by varying it while fixing the shape and appearance vectors for a randomly sampled object. (c) shows an example of novel view synthesis for an observed 2D image.} 
\label{fig:Carla_dis_MCMC}
\end{figure*}

We show more synthesis results for the NeRF-LEBM with amortized inference on the Carla datasets in Figure~\ref{fig:carla_gen}, which are similar to Figure~\ref{fig:Carla_different_view}. We can see those sampled cars are meaningful and their multi-view synthesis results are consistent. Then we show the generative results for the NeRF-LEBM with MCMC inference on the Carla dataset in Figure~\ref{fig:Carla_dis_MCMC}. The model can also correctly disentangle the appearance, shape and camera pose.   

\subsection{More results for the experiment in Section~\ref{sec:unseen_obj} Inferring 3D structures of unseen 2D objects}
\begin{figure}[h!]
    \centering
    \includegraphics[width=.45\textwidth]{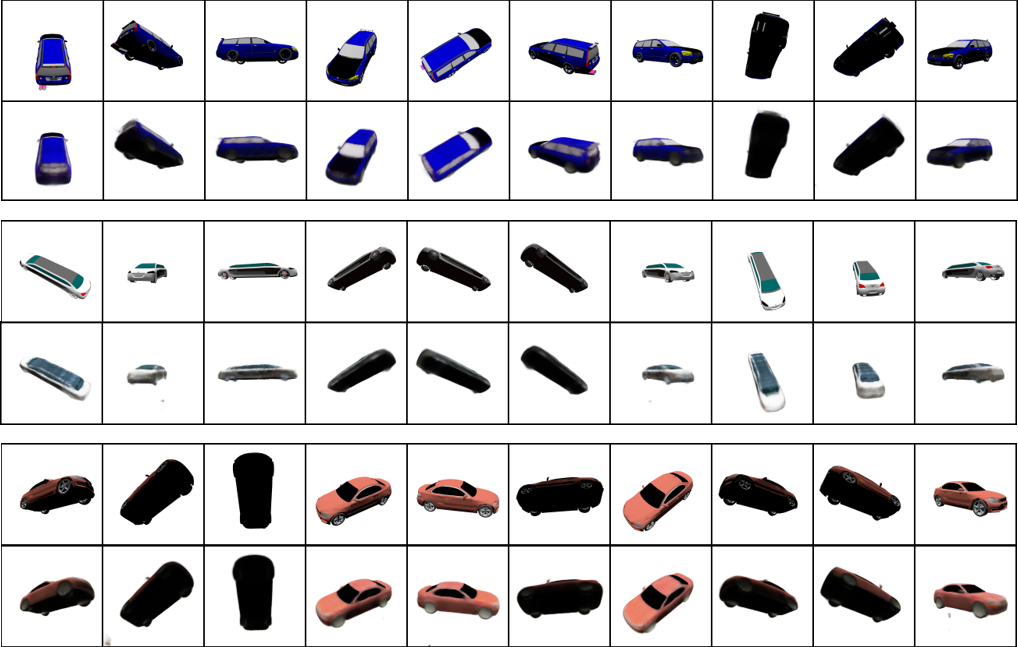}
    \caption{Novel view synthesis results for objects in the ShapeNet Car training set. Each panel shows one case, in which the first row shows the ground truth images while the second row displays the novel view synthesis results by our model.}
    \label{fig:deonly_train_recons}
\end{figure}

In the main paper, we show two-shot novel view synthesis results on the ShapeNet Car testing set in Figure~\ref{fig:SRN_2_shot}. Here in Figure~\ref{fig:deonly_train_recons}, we show the novel view synthesis results for cars in training image. More specifically, for each object in the training set, we use the inferred $\textbf{z}^a$ and $\textbf{z}^s$ from the seen views and synthesize unseen views that hasn't been used during training. Figure~\ref{fig:deonly_train_recons} presents three cases. For each case, the first row shows the ground truth images, and the second row displays the synthesized results by our model.
Our model can correctly output unseen views when camera poses are given.  

\subsection{More results for the experiment in Section~\ref{sec:no_pose} Learning with unknown camera poses}
\begin{figure*}[h]
\centering
\begin{subfigure}{.32\linewidth}
    \centering
    \includegraphics[width=0.99\textwidth]{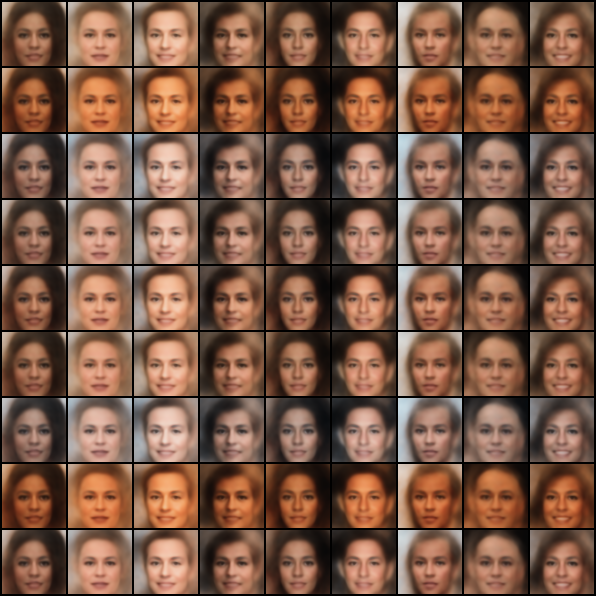}
    \caption{}\label{fig:unsup_interp_celeba}
\end{subfigure}
\hspace{-0.2em}
\begin{subfigure}{.32\linewidth}
    \centering
    \includegraphics[width=0.99\textwidth]{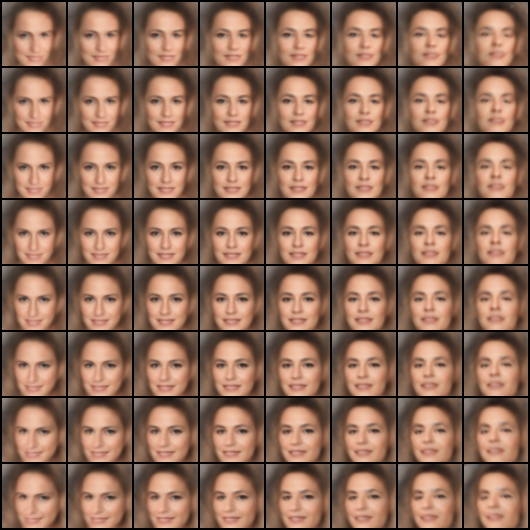}
    \caption{}
    \label{fig:unsup_pose_celeba}
\end{subfigure}
\hspace{-0.2em}
\begin{subfigure}{.32\linewidth}
    \centering
    \includegraphics[width=0.99\textwidth]{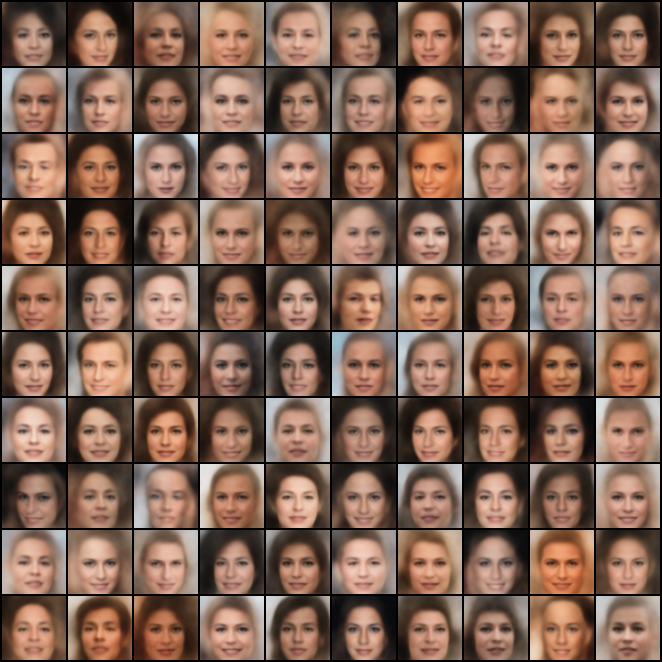}
    \caption{} \label{fig:unsup_syn_celeba}
\end{subfigure}

\caption{Synthesis results on $64 \times 64$ CelebA dataset without knowing camera poses. (a) The objects in each row share the same appearance vector $\textbf{z}^a$ but have different shape vectors $\textbf{z}^s$ while the objects in each column share the same $\textbf{z}^s$ but have different $\textbf{z}^a$. They all share the same camera pose $\boldsymbol{\xi}$. (b) The Effect of changing the camera pose while fixing the shape and appearance vectors for a sampled object.  (c) Generated samples by randomly sampling $\textbf{z}^a,$ $\textbf{z}^s$ and camera pose $\boldsymbol{\xi}$.}
\label{fig:unsup_celeba}
\end{figure*}

\begin{figure}
    \centering
    \includegraphics[width=.32\textwidth]{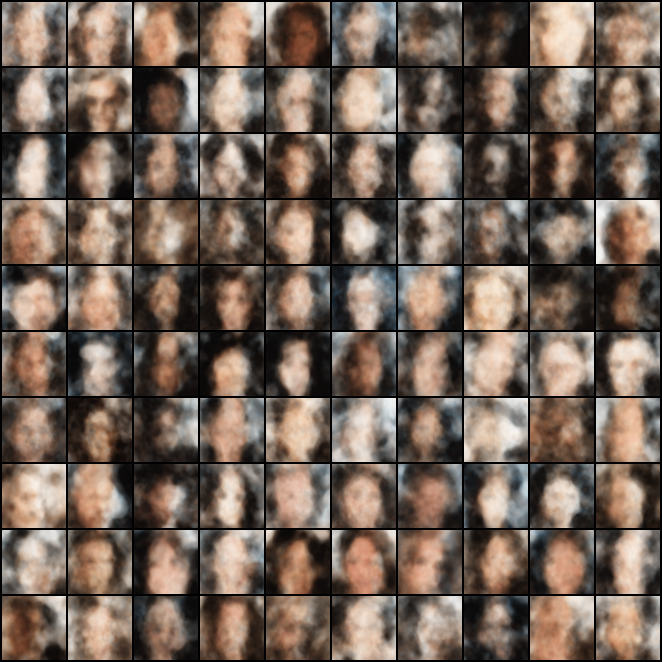}
    \caption{Images generated by a VAE baseline (Gaussian prior), trained on the CelebA dataset without camera poses.}
    \label{fig:celeba_gaussian_prior}
\end{figure}

In the main paper, we show the results on the synthetic dataset, Carla. We also work on a real world dataset. The CelebA dataset~\cite{liu2015faceattributes} contains 200k face images. Unlike the synthetic datasets, this dataset does not provide any camera pose information. We learn our model from this dataset without knowing the camera poses. We preprocess the face images by cropping a $128\times128$ patch from the center of each image and reshaping it into a $64 \times 64$ image. We show the results on CelebA dataset in Figure~\ref{fig:unsup_celeba}. We further show the generated examples obtained from a VAE baseline with a simple Gaussian prior in Figure~\ref{fig:celeba_gaussian_prior}. As we can see, our model can learn to generate meaningful samples without using the ground truth camera poses and the EBM prior provides much better generation results than the simple Gaussian prior.  

\end{document}